\documentclass[12pt]{article}
\usepackage{arxiv}

\usepackage[utf8]{inputenc} 
\usepackage[T1]{fontenc}    
\usepackage{hyperref}       
\usepackage{url}            
\usepackage{booktabs}       
\usepackage{amsfonts}       
\usepackage{nicefrac}       
\usepackage{microtype}      
\usepackage{lipsum}
\usepackage{graphicx}
\graphicspath{ {./images/} }

\title{Can Artificial Intelligence Reconstruct Ancient Mosaics?}

\author{
  Fernando Moral-Andr\'es\ \\
  Universidad Antonio de Nebrija\\
  28015 Madrid, Spain \\
  \texttt{fmoral@nebrija.es} \\
   \And
  Elena Merino-G\'omez \\
  Universidad de Valladolid \\
  47011 Valladolid, Spain \\
  \texttt{elena.merino.gomez@uva.es} \\
   \And
  Pedro Reviriego \\
  Universidad Polit\'ecnica de Madrid\\
  28040 Madrid, Spain \\
  \texttt{pedro.reviriego@upm.es} \\
  \And
  Fabrizio Lombardi \\
  Northeastern University\\
  02115 Boston, US \\
  \texttt{lombardi@ece.neu.edu} \\
}

\begin{document}
\maketitle
\begin{abstract}

A large number of ancient mosaics have not reached us because they have been destroyed by erosion, earthquakes, looting or even used as materials in newer construction. To make things worse, among the small fraction of mosaics that we have been able to recover, many are damaged or incomplete. Therefore, restoration and reconstruction of mosaics play a fundamental role to preserve cultural heritage and to understand the role of mosaics in ancient cultures. This reconstruction has traditionally been done manually and more recently using computer graphics programs but always by humans. In the last years, Artificial Intelligence (AI) has made impressive progress in the generation of images from text descriptions and reference images. State of the art AI tools such as DALL-E2 can generate high quality images from text prompts and can take a reference image to guide the process. In august 2022, DALL-E2 launched a new feature called outpainting that takes as input an incomplete image and a text prompt and then generates a complete image filling the missing parts. In this paper, we explore whether this innovative technology can be used to reconstruct mosaics with missing parts. Hence a set of ancient mosaics have been used and reconstructed using DALL-E2; results are promising showing that AI is able to interpret the key features of the mosaics and is able to produce reconstructions that capture the essence of the scene. However, in some cases AI fails to reproduce some details, geometric forms or introduces elements that are not consistent with the rest of the mosaic. This suggests that as AI image generation technology matures in the next few years, it could be a valuable  tool for mosaic reconstruction going forward.

\end{abstract}



\section{Introduction}

The cultural heritage that we have from ancient civilizations is only partial because many artworks have been lost or damaged for example due to erosion, earthquakes \cite{Earthquakes}, looting or even used as materials in newer constructions. When referring to bidimensional art pieces, sometimes only fragments of a painting or a mosaic have been recovered, but the missing parts can be reconstructed based on the fragments available. Due to their discontinuous nature and their fragility, mosaics are artistic objects prone to damage and incompleteness. Thousands of examples have survived only in a fragmentary state, providing an excellent field for experimentation on image completion. This has been traditionally performed manually and more recently using software tools \cite{virtualrestoration}. Digital image processing techniques play a fundamental role in the preservation and restoration of cultural heritage \cite{digitalimagingcultural} in general and of ancient mosaics in particular \cite{DigitalMosaicReconstructions1},\cite{DigitalMosaicReconstructions2}. These image processing tools greatly facilitate the work of researchers for analyzing mosaics and their restoration but the process is still human guided as for example, the graphical design using software tools. 

More recently, the use of Artificial Intelligence (AI) has been proposed for different tasks related to cultural heritage, for example to manage cultural data and aid visitors to interpret museums \cite{CulturalAI} or to classify elements in models of historical buildings using sophisticated decision tree algorithms \cite{AI-BIM}. In the case of mosaics, genetic algorithms \cite{ImageSegmentationforMosaics} and deep learning \cite{ImageSegmentationML} have been used to segment mosaics; as AI progresses, it will find new applications in the preservation of cultural heritage. 

One area that has seen tremendous progress in the last few years is AI based image generation from text. Many text-to-image AI generators are available, such as for example Imagen \cite{Imagen} or Parti \cite{Parti} from Google, Cogview \cite{Cogview}, Midjourney\footnote{Available at \url{www.midjourney.com}} or Stability AI\footnote{Available at \url{https://stability.ai/blog/stable-diffusion-announcement}}. The most famous generator is DALL-E from OpenAI \cite{DALLE1},\cite{DALLE2} that has been, for example, used recently to create the font cover of a well-known magazine\footnote{Available at \url{www.cosmopolitan.com/lifestyle/a40314356/dall-e-2-artificial-intelligence-cover/}}. These tools take as input a prompt text and optionally a reference image to produce a new image that reflects the description in the text conditioned to the reference image. To do so, the generators have billions of parameters, and huge datasets are used for training. For example, the latest Google generator, Parti has been trained on several billions of text/image pairs and can have up to 20 billion parameters \cite{Parti}. This complexity pays off as the tools provide high quality images for many text inputs.

In August 2022, DALL-E introduced a new feature called "outpainting" that takes as input an incomplete image and fills in the missing parts\footnote{\url{https://openai.com/blog/dall-e-introducing-outpainting/}}. This fits exactly the needs of mosaic reconstruction. In this paper, we explore for the first time the use of a state-of-the-art AI image generation tool (DALL-E) to perform the virtual reconstruction of ancient mosaics. The rest of the paper is organized as follows. In section 2, we study the reconstruction of damaged mosaics evaluating DALL-E performance on figurative and geometric mosaics. Then in section 3, we take mosaics that are well preserved and artificially remove parts of the image to then perform reconstruction with DALL-E. This enables us to compare the reconstruction against the original image. The results and potential avenues for further research are discussed in section 4. The paper ends with the conclusion in section 5.

\section{Initial Evaluation of AI Mosaic Reconstruction}
\label{sec:evaluation}

To evaluate the use of AI for the reconstruction of ancient mosaics, a set of eight mosaics crafted with the techniques of opus tesselatum and opus vermiculatum \cite{HandbookRomanArt}, has been used. Most of them come from the Jean Paul Getty collection and high-quality images of the mosaics are available through the Open Content Program of the Getty Foundation\footnote{\url{https://www.getty.edu/projects/open-content-program/}}. The rest are taken from the British Museum and the Louvre Museum, and are also publicly available. The set includes the most common issues on ancient mosaics such as figurative scenes with humans and animals as well as geometric mosaics with different levels of missing fragments. 

For each image, reconstruction has been performed using DALL-E2 \cite{DALLE2} outpainting feature; this feature allows to complete images with missing fragments on a window of 1024 by 1024 pixels. Depending on the aspect ratio and resolution of the original image, we either adjust its resolution to fit in such a window, or perform the reconstruction by running outpainting on several parts of the original image; in all cases, the simple text prompt "Roman mosaic" was used in conjunction with the incomplete image. The results for each mosaic are presented and discussed in the following subsections. 

\subsection{Mosaic Floor with Achilles and Briseis}

The first mosaic represents a scene from Homer’s Iliad. In more detail, the mosaic shows Briseis being taken away from Achilles by Talthybios and Eurybates to be given to Agamemnon. The mosaic is thought to come from Antioch (present-day Antakya, Turkey) and to correspond to the AD. 100-300 period\footnote{Further details as well as the image with different resolutions are available at \url{https://www.getty.edu/art/collection/object/103SPQ}}. From a reconstruction point of view, relevant parts of the image are missing.

The original image and the AI reconstructed image are shown in Figure 1; the AI tool is able to reconstruct the scene. However, the quality of the reconstructed faces is low and in particular Briseis who should be looking left is reconstructed looking at the front. The herald to its right is reconstructed with a blurry face and his arm that should be holding Brisseis is pointing to one of his elbows. These mistakes in the reconstruction can be partly attributed to the AI tool not having the context of the scene been represented with such level of detail. However, the reconstruction of the faces and some details seem to be genuine limitations of the tool.

\begin{figure*}
   \label{FigM1}
    \centering
      \includegraphics[scale=1.6 ]{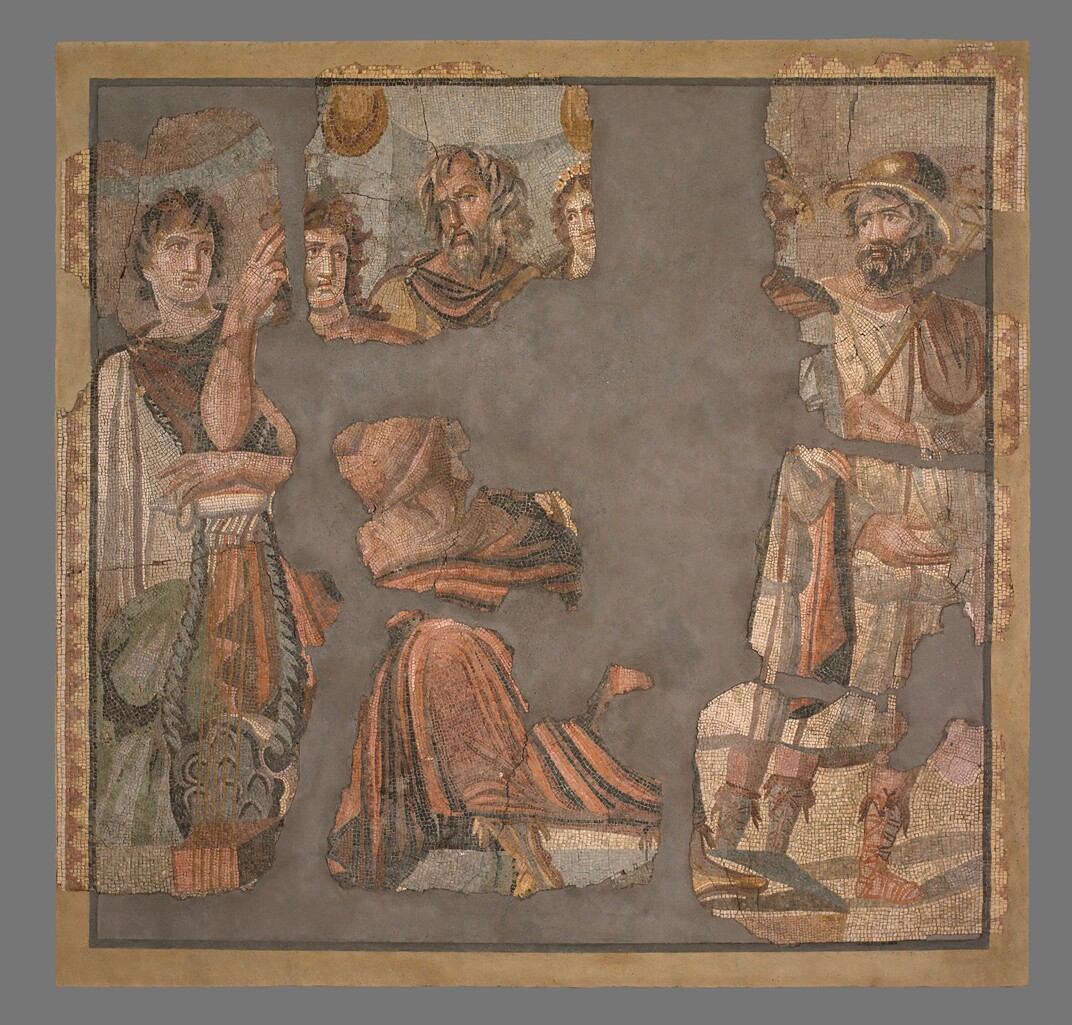}\\
      .\\
      \includegraphics[scale=0.3]{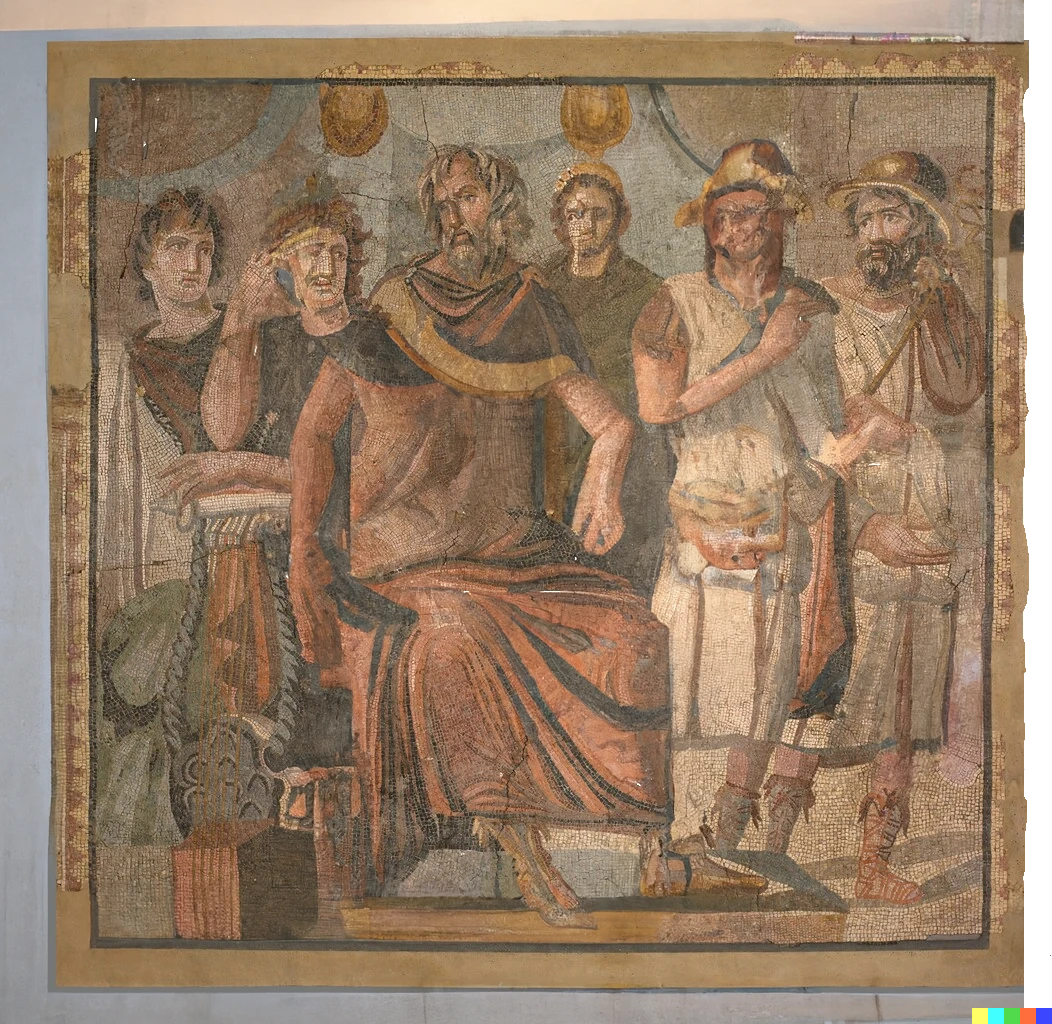}\\
      
    \caption{Achilles and Briseis: original (top), reconstructed (bottom)}
\end{figure*}

\subsection{Amazon Battle}

The second mosaic depicts the battle of an amazon warrior being seized by her cap by a warrior\footnote{Further details as well as the image with different resolutions are available at \url{https://es.wikipedia.org/wiki/Archivo:Amazonomachy_Antioch_Louvre_Ma3457.jpg}}. It is part of the collection at the Louvre Museum; dated to the 2nd half of the 4th century AD, it was found in the excavations in Antioch as the first mosaic. A large fragment of the right side is missing.  

The original image and the AI reconstructed image are shown in Figure 2. In this case, the AI tool is capable to reconstruct the scene, this time with better quality. This may be because the missing part in this mosaic is easier to reconstruct than in the first one; however, even in this mosaic there are errors like adding a fifth leg to the warrior’s horse. 

\begin{figure*}
   \label{FigM2}
    \centering
      \includegraphics[scale=0.22 ]{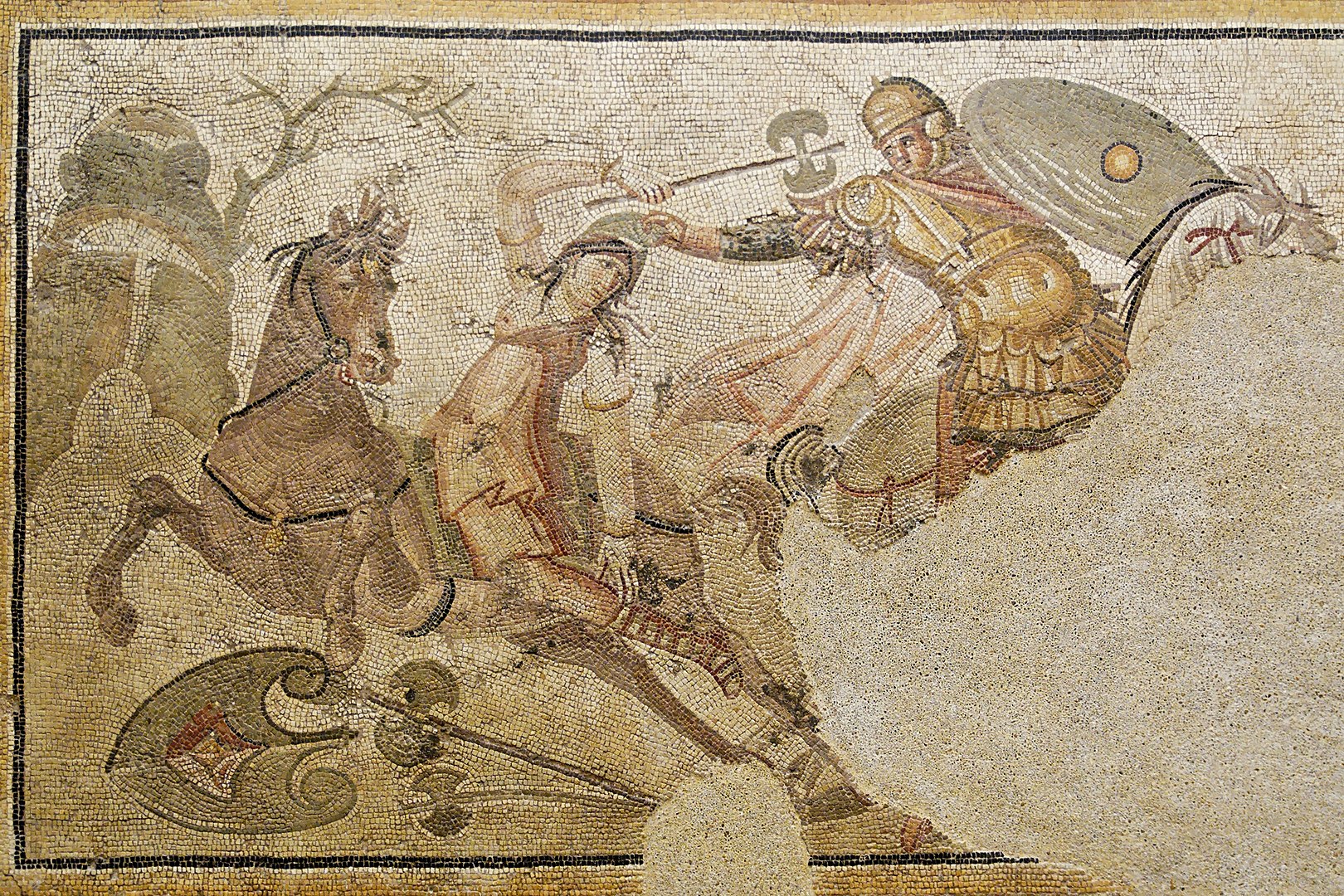}\\
      .\\
      \includegraphics[scale=0.2]{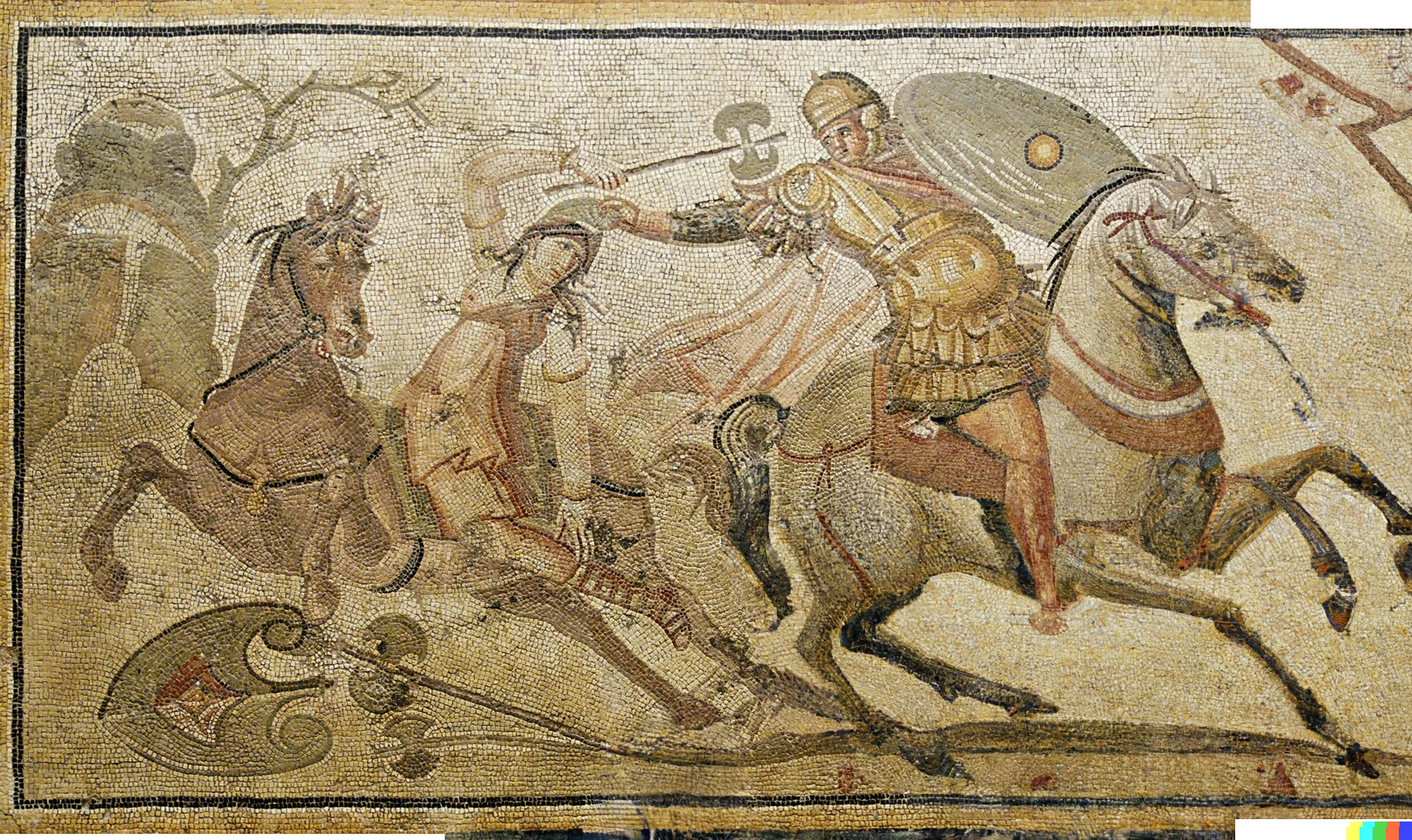}\\
      
    \caption{Amazon Battle: original (top), reconstructed (bottom)}
\end{figure*}

\subsection{Alexander Mosaic}

The third example is a scene from the Battle of Issus that took place in 333 BC between Alexander the Great and Darius III of Persia. The mosaic was found in the House of the Faun in Pompeii and is now part of the collection of the National Archaeological Museum in Naples\footnote{Further details as well as the image with different resolutions are available at \url{https://commons.wikimedia.org/wiki/File:Battle_of_Issus_mosaic_-_Museo_Archeologico_Nazionale_-_Naples_2013-05-16_16-25-06_BW.jpg}}. A significant part of the scene around the figure of Alexander the Great is missing. 

The original image and the AI reconstructed image are shown in Figure 3. The tool is capable to interpret the overall scene but fails on many details. For example, Alexander the Great is riding a rampant horse but it is reconstructed as a standing horse whose back does not look much like a horse. The recreated warriors are standing while they were more likely to be ridding horses as Alexander. The two seals added also look like a bicycle that a warrior is riding, but this may be a coincidence. In this mosaic, some of the errors can be also attributed to the lack of context but the failure to identify the rampant horse seems to be an intrinsic limitation of the AI tool.  

\begin{figure*}
   \label{FigM3}
    \centering
      \includegraphics[scale=0.2 ]{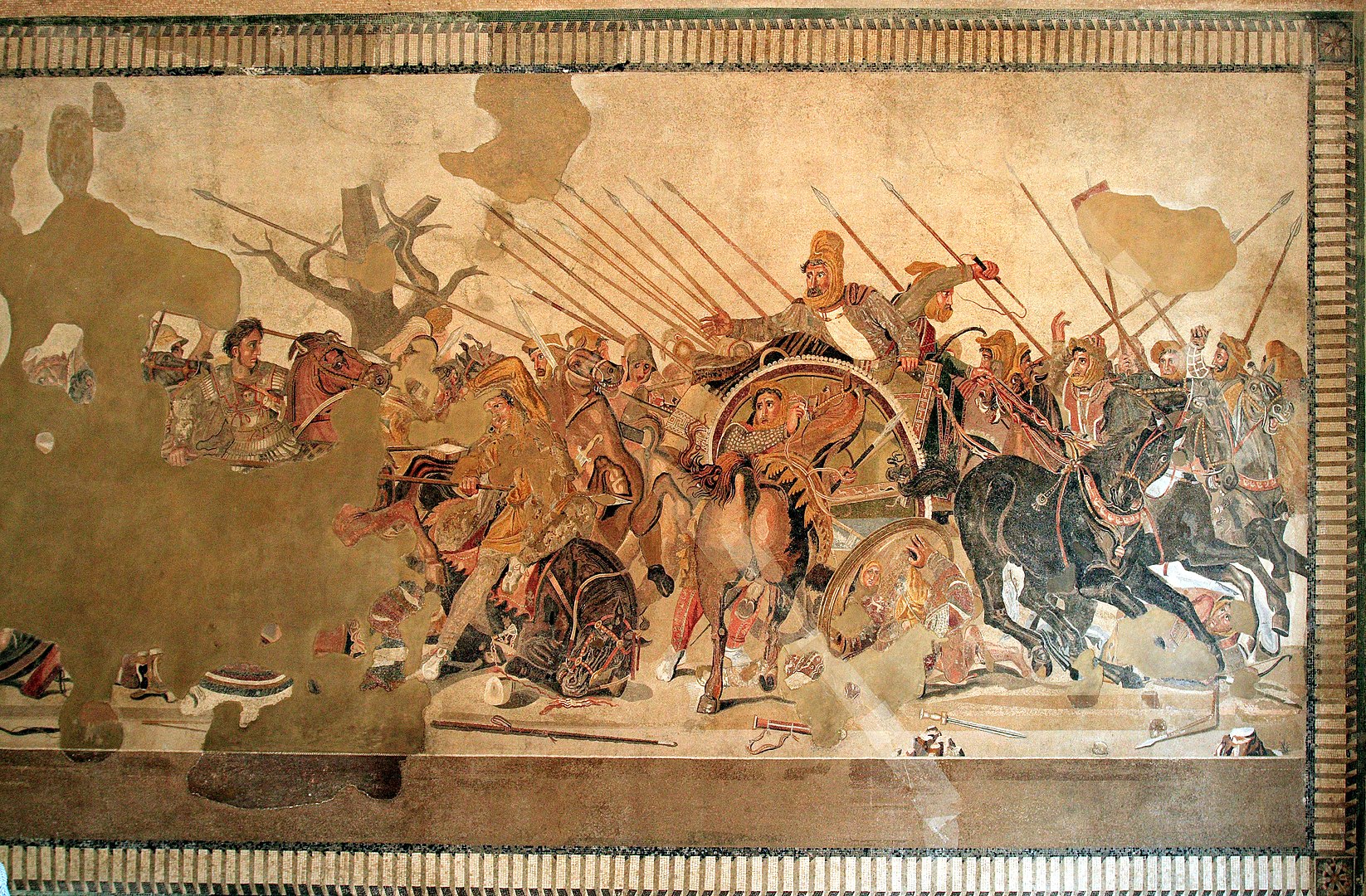}\\
      .\\
      \includegraphics[scale=0.2]{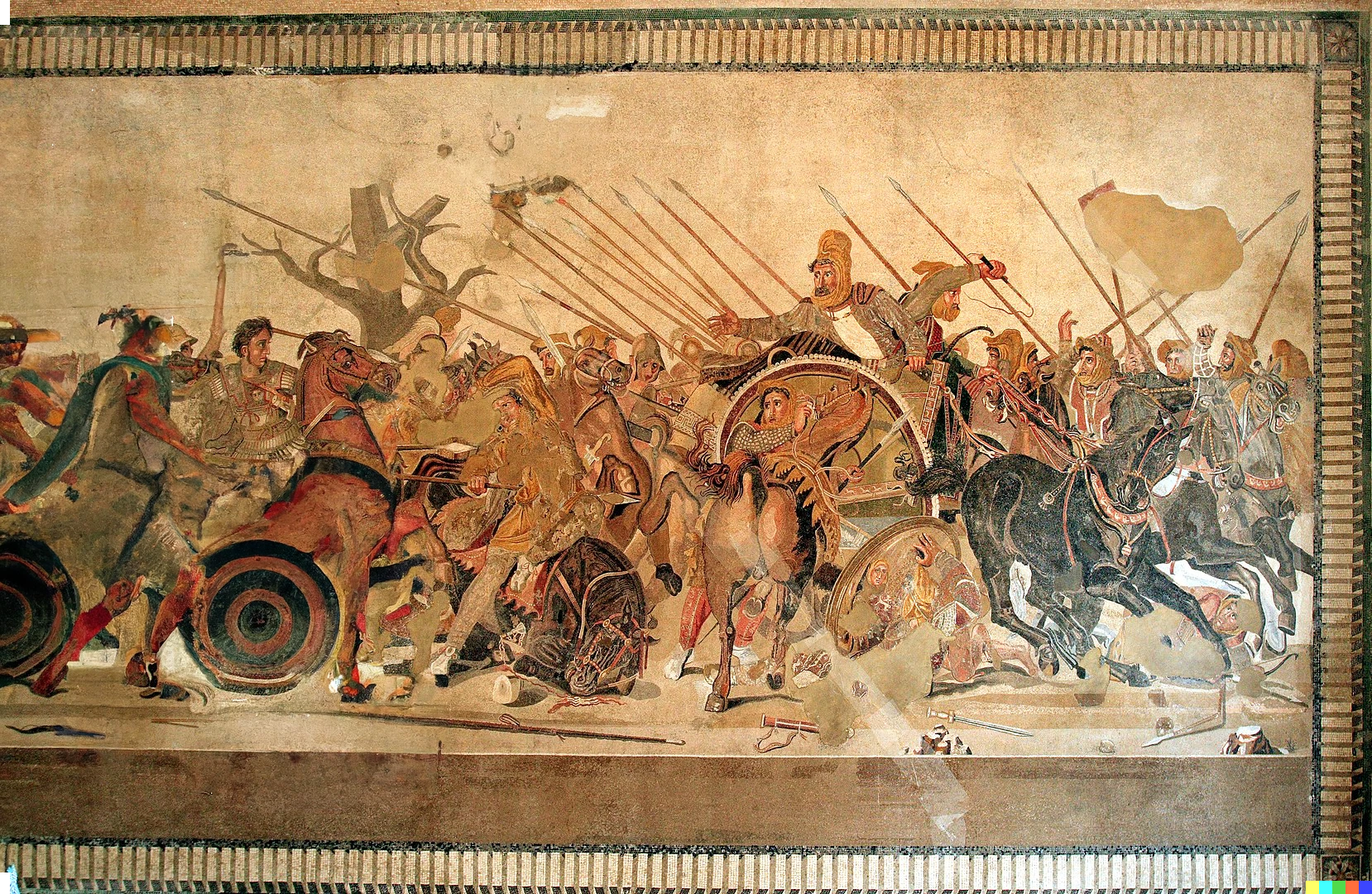}\\
      
    \caption{Alexander in the battle of Issus: original (top), reconstructed (bottom)}
\end{figure*}

\subsection{Lion attacking an Onager}

The fourth mosaic is a scene of a lion attacking an onager that is part of the Getty collection\footnote{Further details as well as the image with different resolutions are available at \url{https://www.getty.edu/art/collection/object/103SYF}}. The mosaic was found in Tunisia and it is thought to correspond to the A.D. 150–200 period. Mosaics of animal fighting were popular in the Roman province of Africa that included Tunisia. In this case, only fragments on the borders of the scene are missing.

The original image and the AI reconstructed image are shown in Figure 4. For this mosaic the AI tool is also able to reconstruct the scene but with errors. The first major error is that instead of recreating the partly missing tree on the right side of the mosaic, a type of flying horse is added. On the bottom of the mosaic, the geometric pattern is only partly recreated, and the colors do not follow the sequence of the rest of the mosaic. 

\begin{figure*}
   \label{FigM4}
    \centering
      \includegraphics[scale=0.58 ]{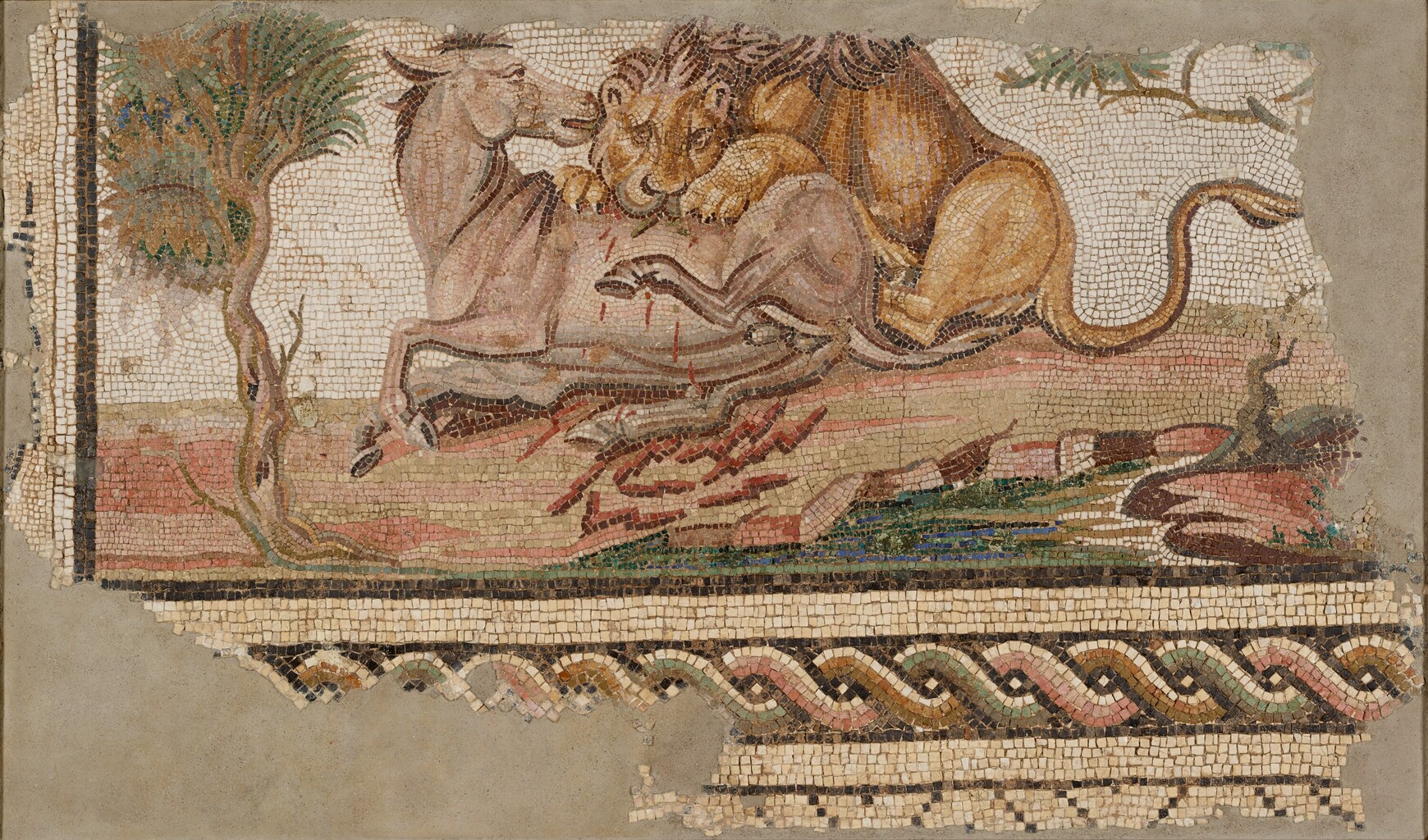}\\
      .\\
      \includegraphics[scale=0.2]{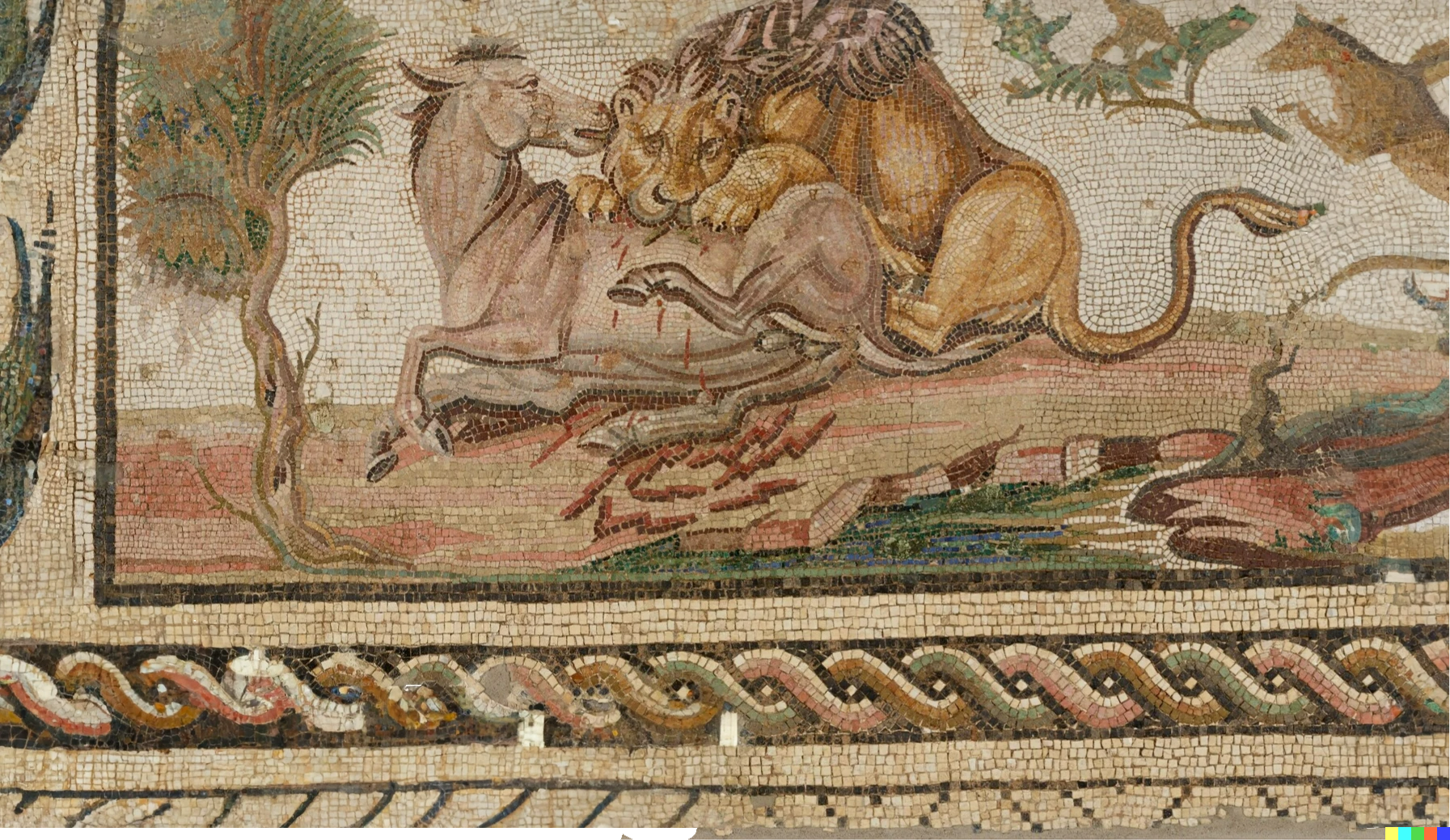}\\
      
    \caption{Lion attacking an onager: original (top), reconstructed (bottom)}
\end{figure*}

\subsection{Two Male Busts}

The fifth mosaic is a panel with two male busts from the 5th century A.D. and is also part of the Getty collection\footnote{Further details as well as the image with different resolutions are available at \url{https://www.getty.edu/art/collection/object/105Y74}}. In this case, only the central part of the mosaic is present, and the rest is missing.

The original image and the AI reconstructed image are shown in Figure 5. The AI tool is capable to reconstruct  the scene; however as a major part of the mosaic is missing, it is arguable whether the reconstruction is correct as would be the case for a manual reconstruction.

\begin{figure*}
   \label{FigM5}
    \centering
      \includegraphics[scale=0.61 ]{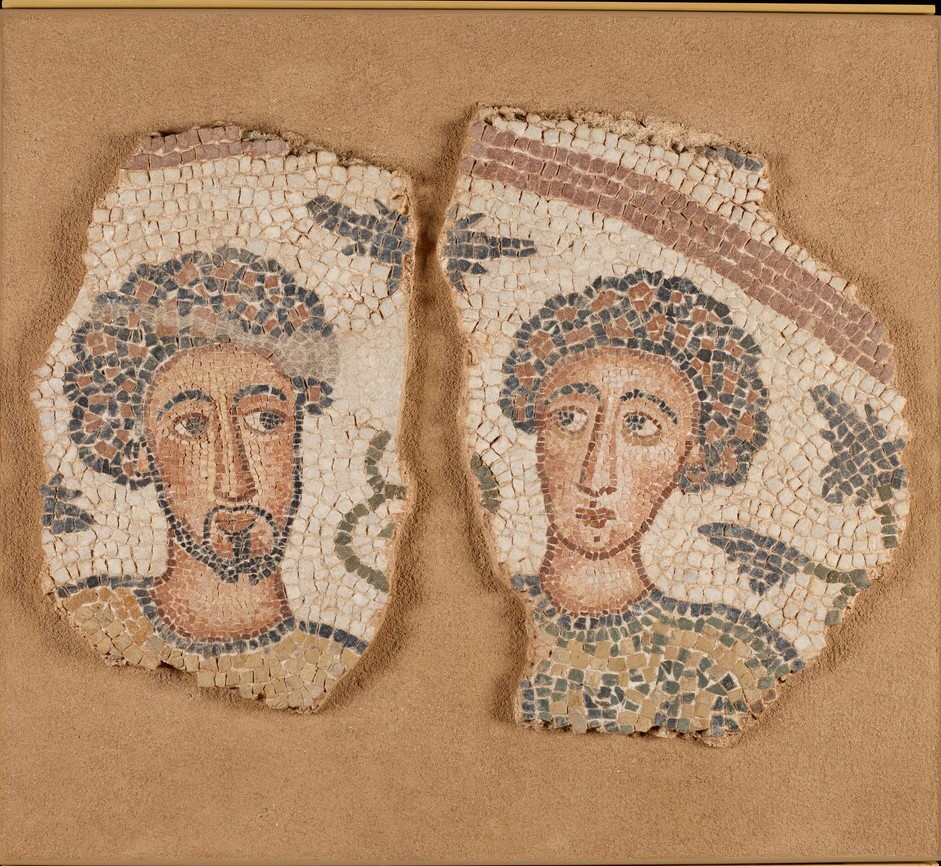}\\
      .\\
      \includegraphics[scale=0.2]{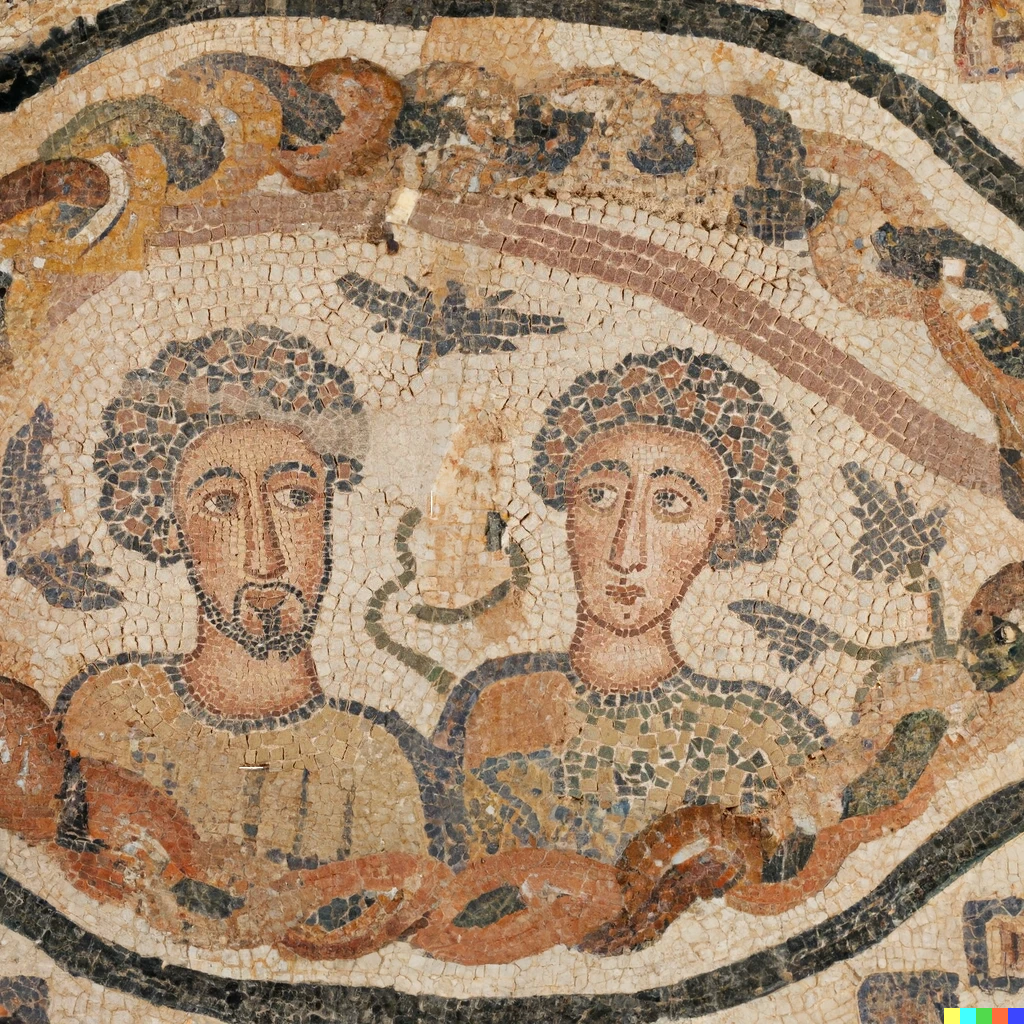}\\
      
    \caption{Two male busts: original (top), reconstructed (bottom)}
\end{figure*}

\subsection{Head of a Season}

The sixth example is a Byzantine mosaic that shows the head of a season and is also part of the Getty collection\footnote{Further details as well as the image with different resolutions are available at \url{https://www.getty.edu/art/collection/object/105X9N}}. In this case, only several scattered parts of the mosaic are missing.

The original image and the AI reconstructed image are shown in Figure 6. In this mosaic, it appears that the AI tool is capable to reconstruct the scene correctly since only small parts of the mosaic are missing, so making the reconstruction easier.

\begin{figure*}
   \label{FigM6}
    \centering
      \includegraphics[scale=0.8 ]{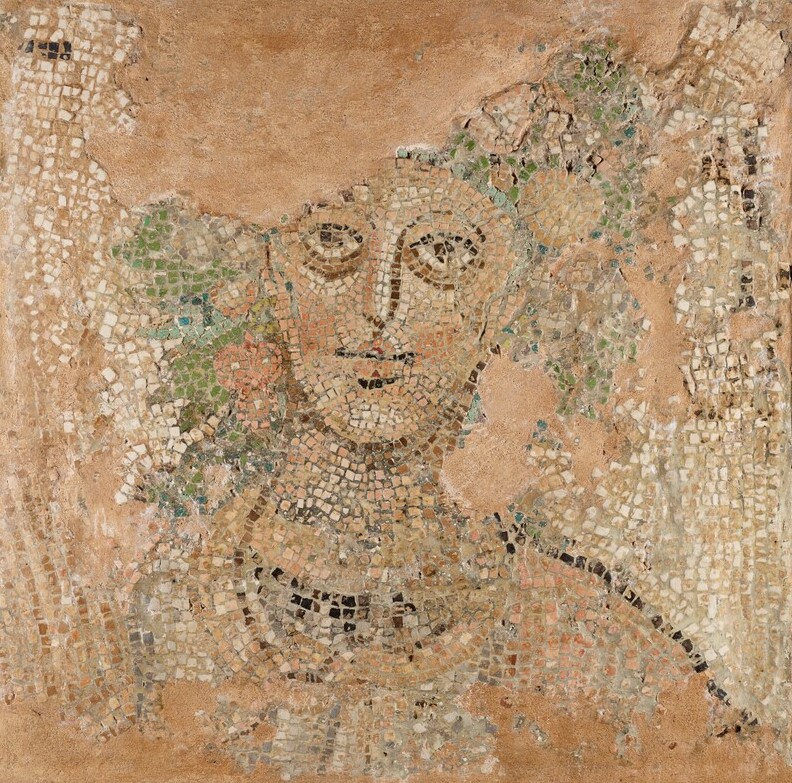}\\
      .\\
      \includegraphics[scale=0.38]{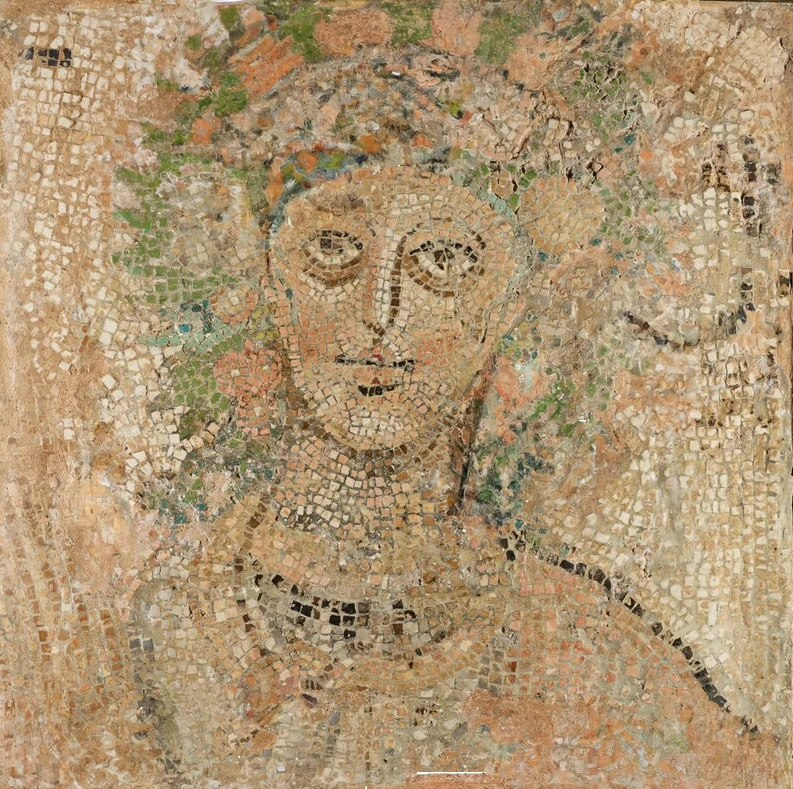}\\
      
    \caption{Head of a Season: original (top), reconstructed (bottom)}
\end{figure*}

\subsection{Floor with Animals}

The two last mosaics are represented by mostly geometric forms rather than figurative scenes. The first one is also part of the Getty collection\footnote{Further details as well as the image with different resolutions are available at \url{https://www.getty.edu/art/collection/object/103SQ4}} and is a floor with geometric forms and small figures of animals. In this case only a small part of the mosaic is missing 

The original image and the AI reconstructed image are shown in Figure 7. The AI tool is capable to reconstruct  the scene correctly interpreting the symmetry and recreate the missing part accordingly. However, it should be noted that it fails to reconstruct a small missing part at the top of the mosaic by inserting a black triangle. 

\begin{figure*}
   \label{FigM7}
    \centering
      \includegraphics[scale=0.62 ]{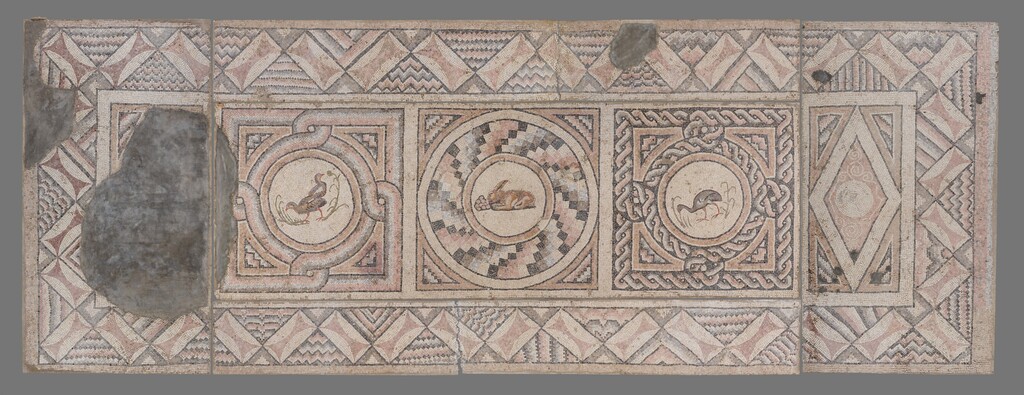}\\
      .\\
      \includegraphics[scale=0.6]{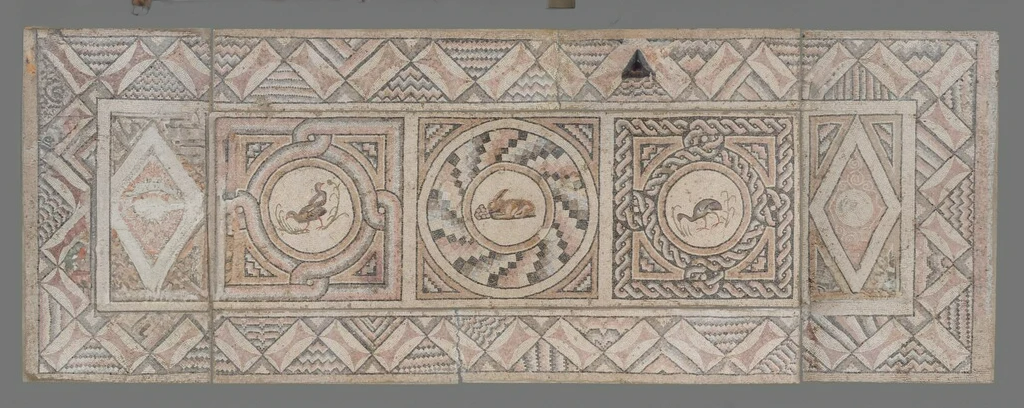}\\
      
    \caption{Floor with animals: original (top), reconstructed (bottom)}
\end{figure*}

\subsection{Thruxton Floor}

The last mosaic is a pavement with geometric forms that is part of the collection of the British Museum\footnote{Further details as well as the image are available at \url{https://www.britishmuseum.org/collection/image/105193001}} and a large part of the mosaic is missing. This mosaic floor when discovered had its central roundel showing a large figure of Bacchus. The heads in the corners probably represent the seasons. The incomplete inscription has not been fully deciphered; it might refer to the owner of the villa \cite{Thruxton}: Quintus, Natalius, Natalinus and Bodeni.

The original image and the AI reconstructed image are shown in Figure 8. The AI tool is capable to reconstruct the main geometric forms but this time the colors used are slightly different from the correct ones. Instead, the tool fails to recreate the central figure of Bacchus; this can be in part attributed to the lack of context, but it is also true that it was common to have figurative scenes in the middle of the mosaics and the tool does not recreate any figure. 

\begin{figure*}
   \label{FigM8}
    \centering
      \includegraphics[scale=1 ]{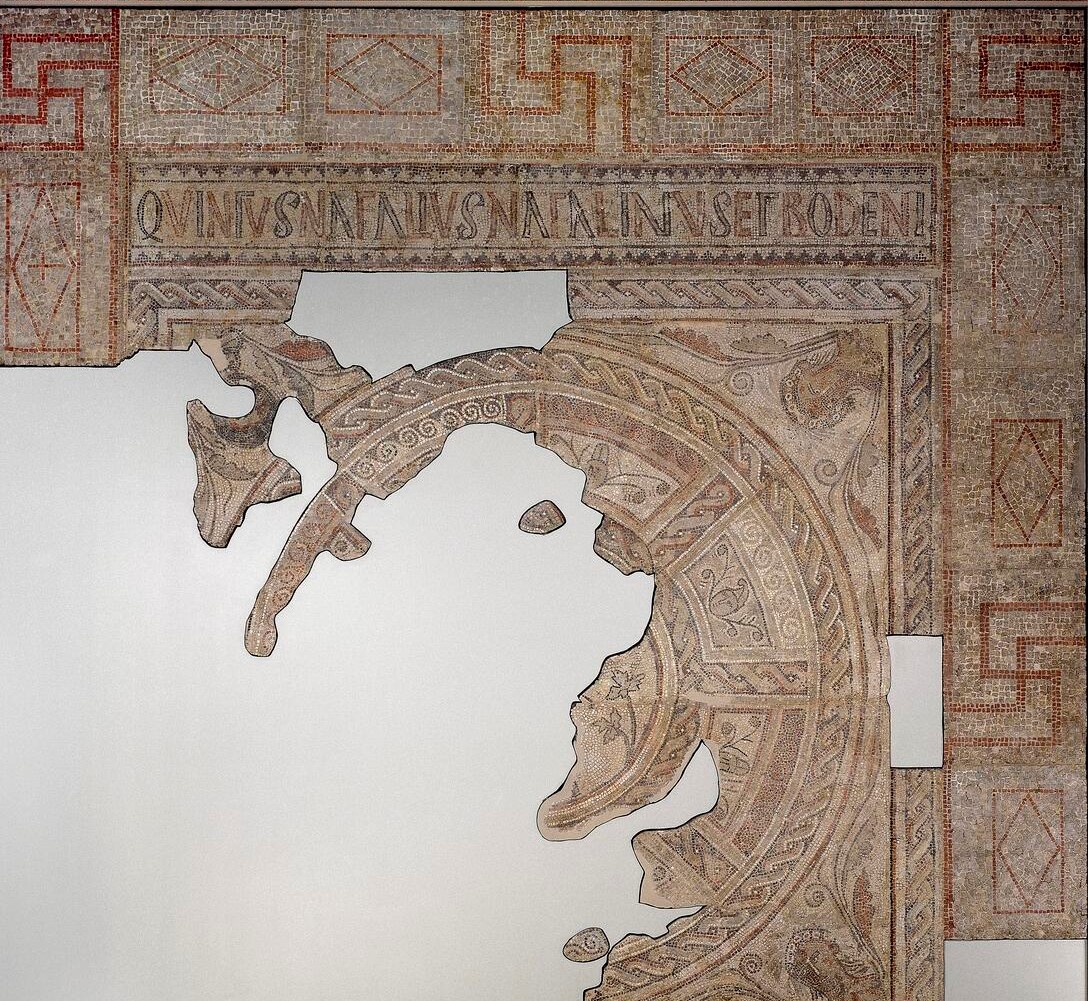}\\
      .\\
      \includegraphics[scale=0.32]{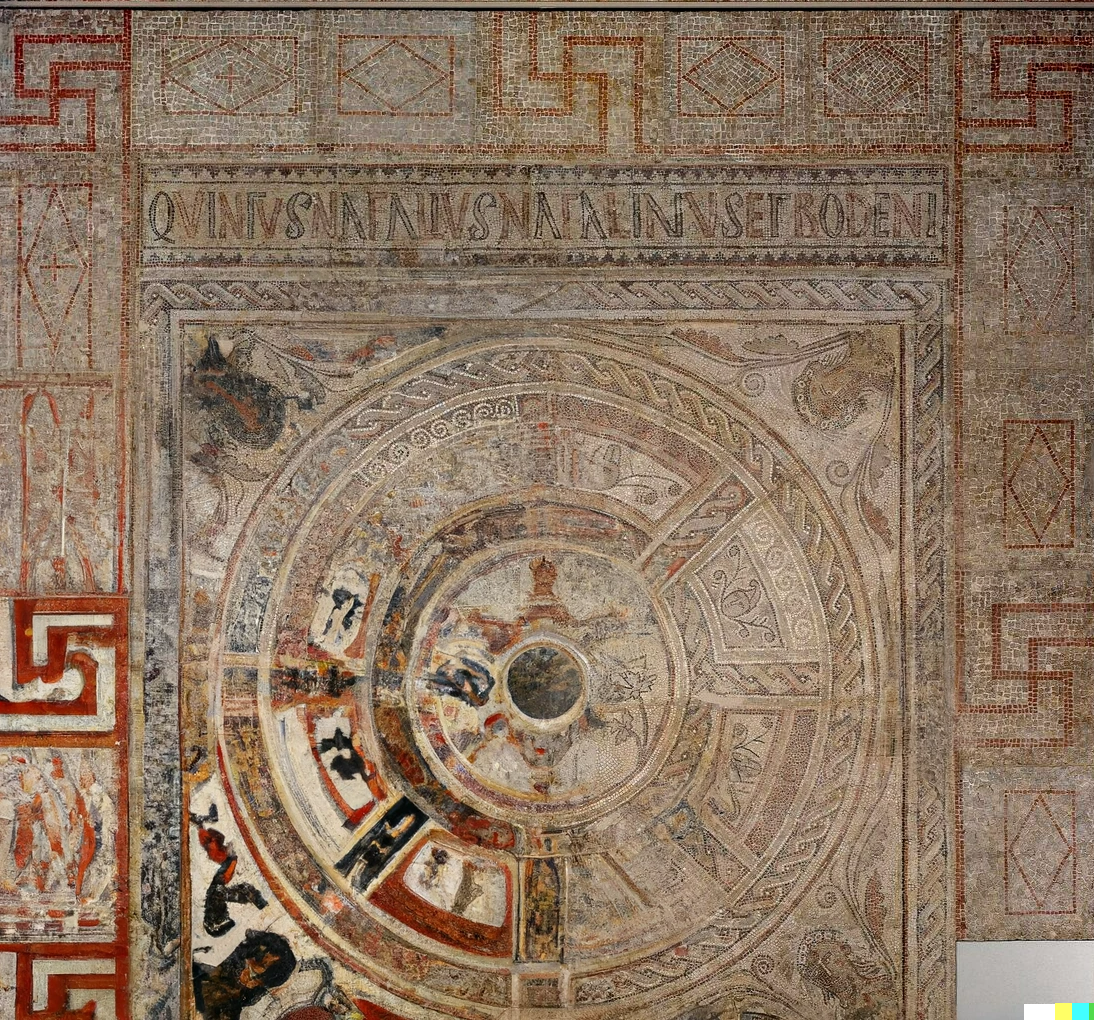}\\
      
    \caption{Thruxton Floor: original (top), reconstructed (bottom)}
\end{figure*}

\section{Emulation of Mosaic Reconstruction}
\label{sec:Emulating}

Instead of a damaged mosaic, an interesting observation is that to evaluate the AI reconstruction capabilities, we can also start from a well-preserved mosaic and then remove parts of the image to test the ability of the AI to perform reconstruction. This allows to evaluate damages on any arbitrary parts of the mosaic as well as to compare with the original image to assess the quality of the reconstruction. Hence, a smaller set of four additional mosaics from the Getty collection have been used; as described in the following sections, we also present the results and analysis of the AI reconstructions.

\subsection{Combat Between Dares and Entellus}

The first mosaic is a scene of the fight between Dares and Entellus; this mosaic was part of a larger floor of a villa in southern France and dated between A.D. 175–200\footnote{Further details as well as the image with different resolutions are available at \url{https://www.getty.edu/art/collection/object/103SQM}}. This fight is part of a passage from Virgil’s Aeneid in which Aeneas honored the anniversary of his father's death by holding elaborate funeral games, including a boxing match. This match pitted the Trojan Dares against the local Sicilian champion Entellus. The image has been altered by removing a large part of the warriors and also part of a corner.

The original image, the editing with the modifications introduced artificially and the AI reconstructed image are shown in Figure 9. The AI tool is capable to reconstruct the scene correctly but puts clothes on the warriors that were originally naked. This may be due to the protection mechanisms that image generation tools typically have to avoid explicit content such as nudity. Another mistake is that the warrior on the right is depicted front facing, instead of rear facing.

\begin{figure*}
   \label{FigM9}
    \centering
      \includegraphics[scale=0.34 ]{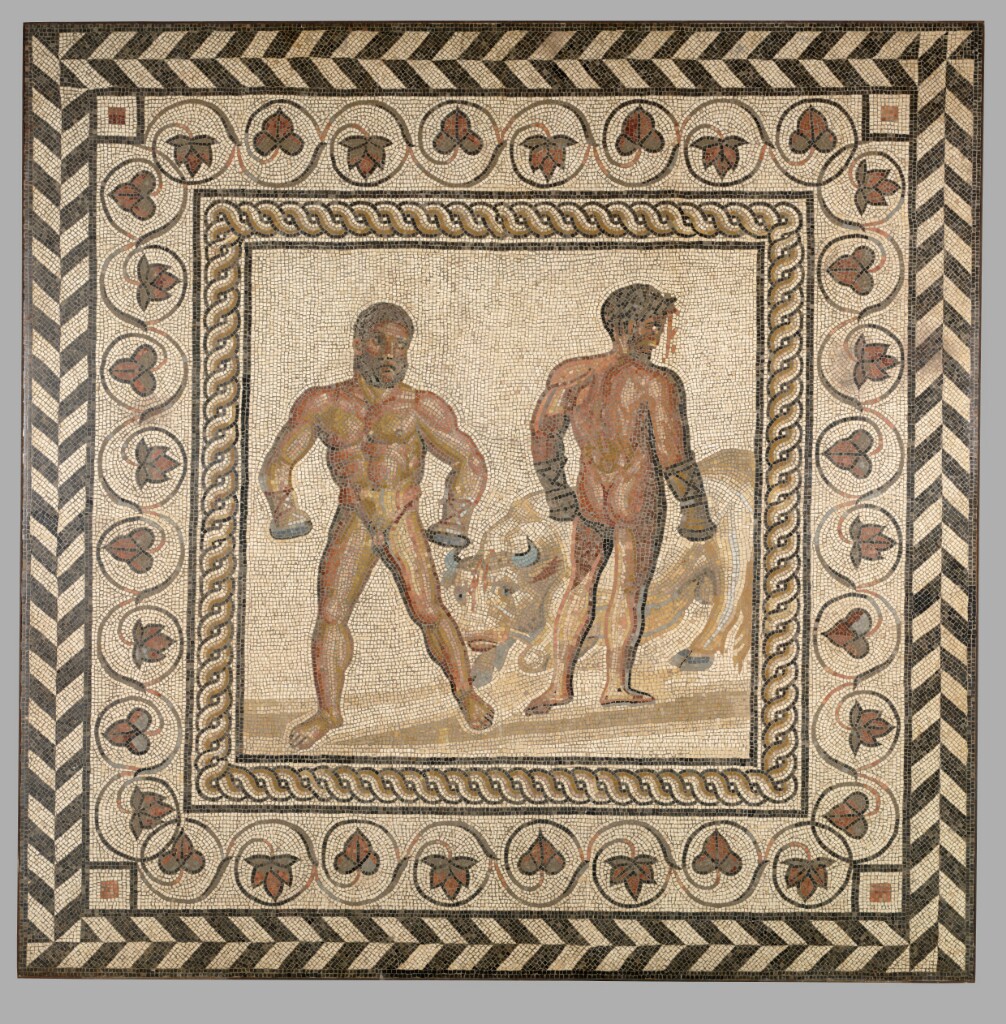}\\
      .\\
      \includegraphics[scale=0.2]{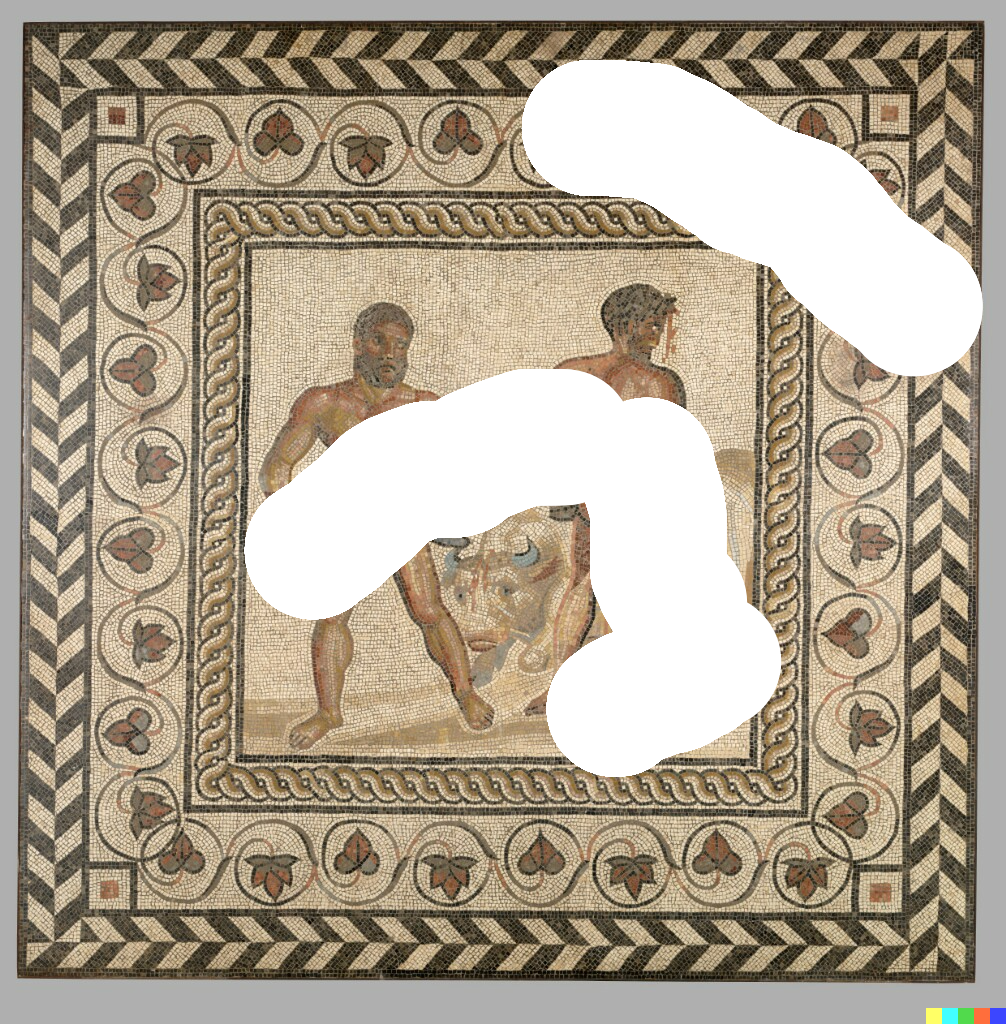}\\
      .\\
      \includegraphics[scale=0.2]{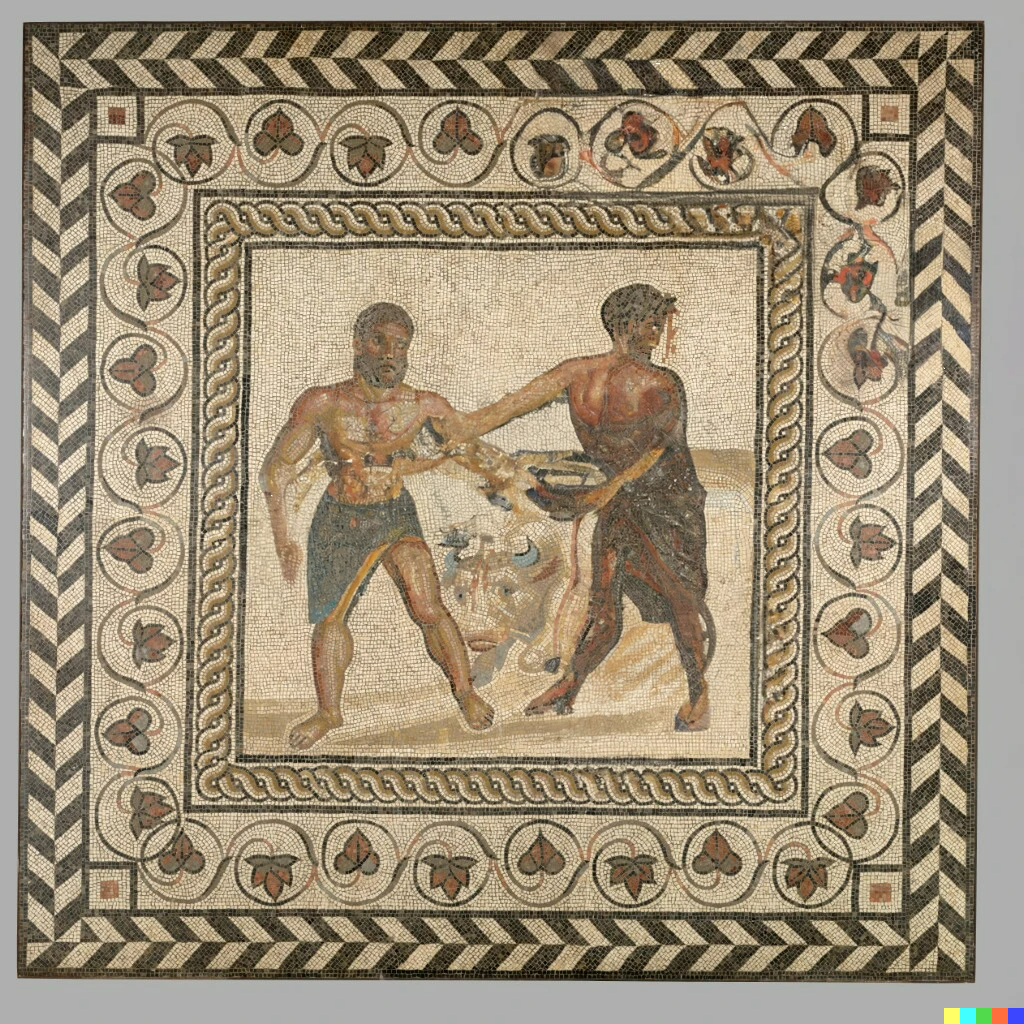}\\
    \caption{Combat Between Dares and Entellus: original (top), emulated damage (middle), reconstructed (bottom)}
\end{figure*}

\subsection{Birds}

The second mosaic depicts two birds, it comes from Rome and is dated in the 3rd or 4th century A.D.\footnote{Further details as well as the image with different resolutions are available at \url{https://www.getty.edu/art/collection/object/105X9J}}. In this case the damage is localized on several small parts of the mosaic including the two birds themselves. 

The original image, the editing with the damage introduced artificially and the AI reconstructed image are shown in Figure 10. The AI tool is capable to reconstruct the scene correctly but it fails to reconstruct the branches on the bottom of the mosaic. The AI tool considers the missing parts of the crossed branches, below the birds, as a discontinuity in a curve; it reconstructs the entire mosaic as in the Gestalt Principle of Closure, in which incomplete elements, with interruptions or gaps, tend to be mentally reconstructed \cite{Gestaltprinciples}. This has been observed in some AI image classification systems that to some extent tend to follow the Gestalt Principles \cite{GestaltCNN}.

\begin{figure*}
   \label{FigM10}
    \centering
      \includegraphics[scale=0.16 ]{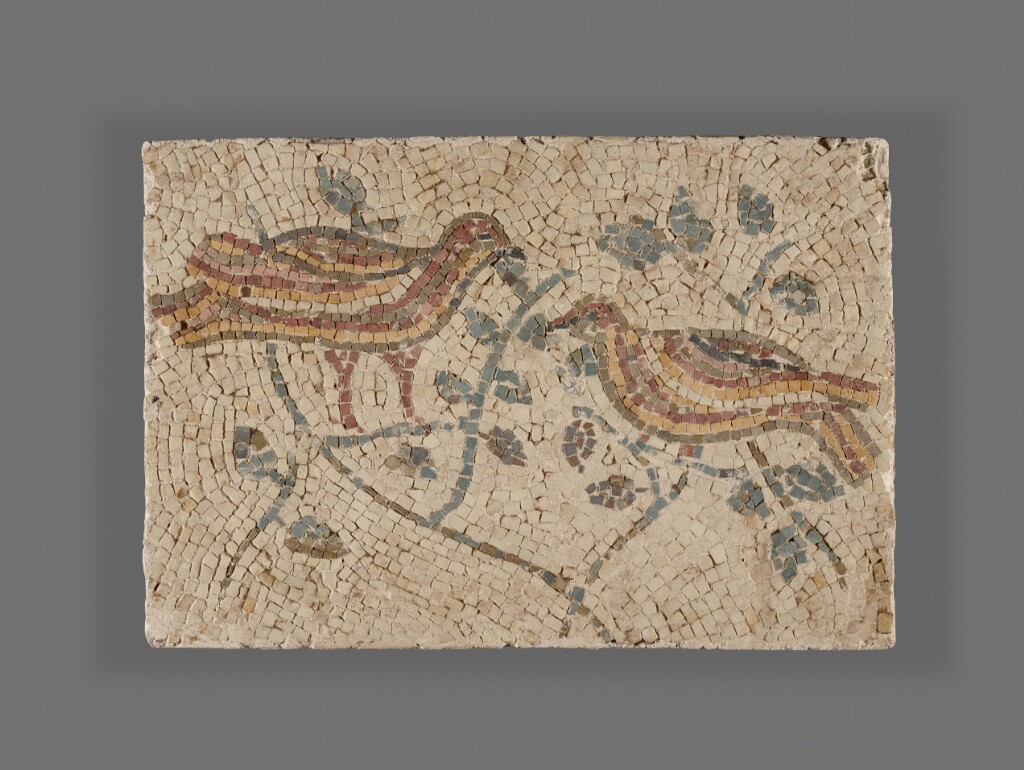}\\
      .\\
      \includegraphics[scale=0.22]{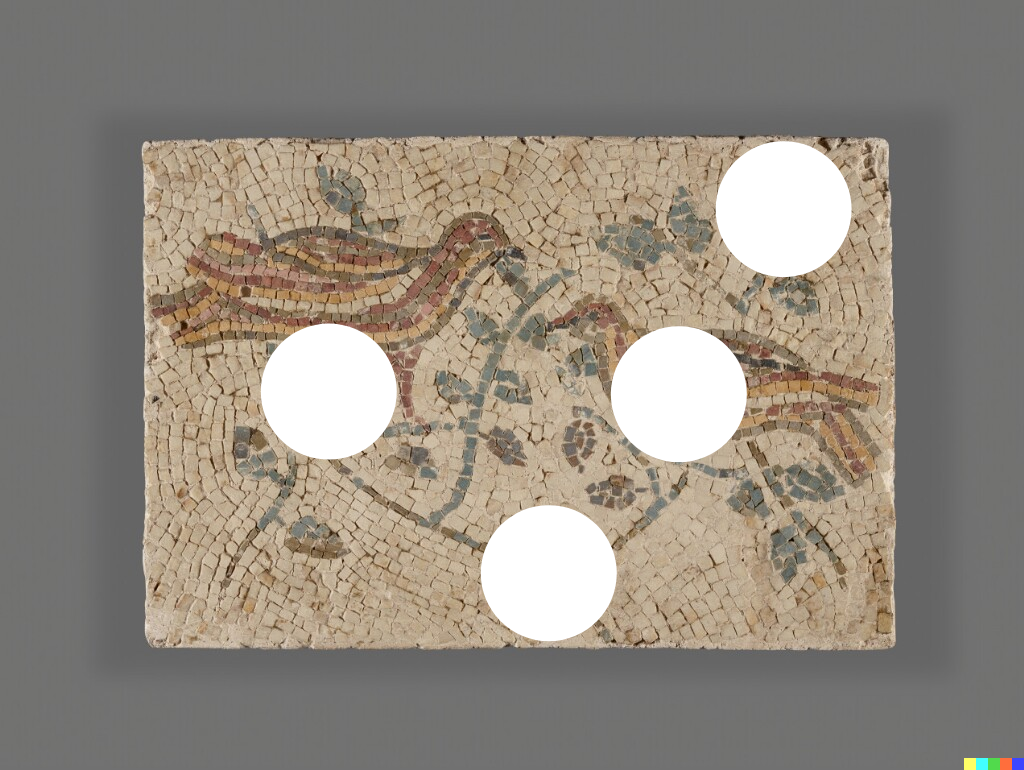}\\
      .\\
      \includegraphics[scale=0.3]{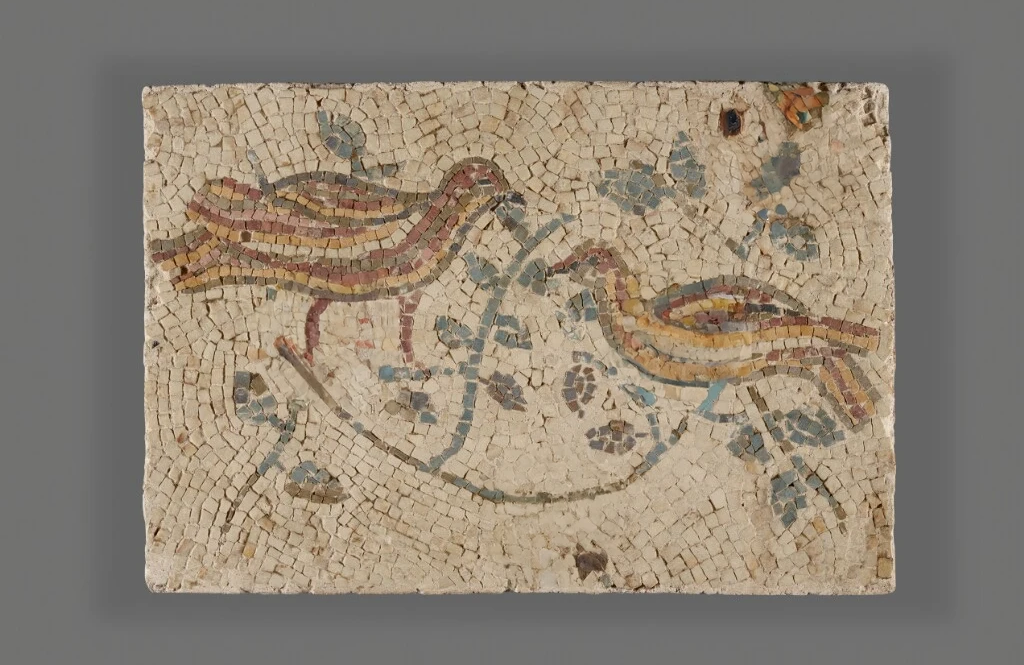}\\
    \caption{Birds: original (top), emulated damage (middle), reconstructed (bottom)}
\end{figure*}

\subsection{Mosaic Floor with Head of Medusa}

The third mosaic is a floor with geometric shapes and the head of a medusa in the center, from about A.D. 115–150 in Rome \footnote{Further details as well as the image with different resolutions are available at \url{https://www.getty.edu/art/collection/object/103SQK}}. Extensive damage has been made to the bottom-right part of the mosaic.

The original image, the editing with the damage introduced artificially and the AI reconstructed image are shown in Figure 11. The AI tool is capable of extrapolating the complex geometric shapes from other parts of the mosaic and produce an almost perfect reconstruction. However, the kylix on the bottom right corner is slightly different from the other three while it should be exactly the same.

\begin{figure*}
   \label{FigM11}
    \centering
      \includegraphics[scale=0.21 ]{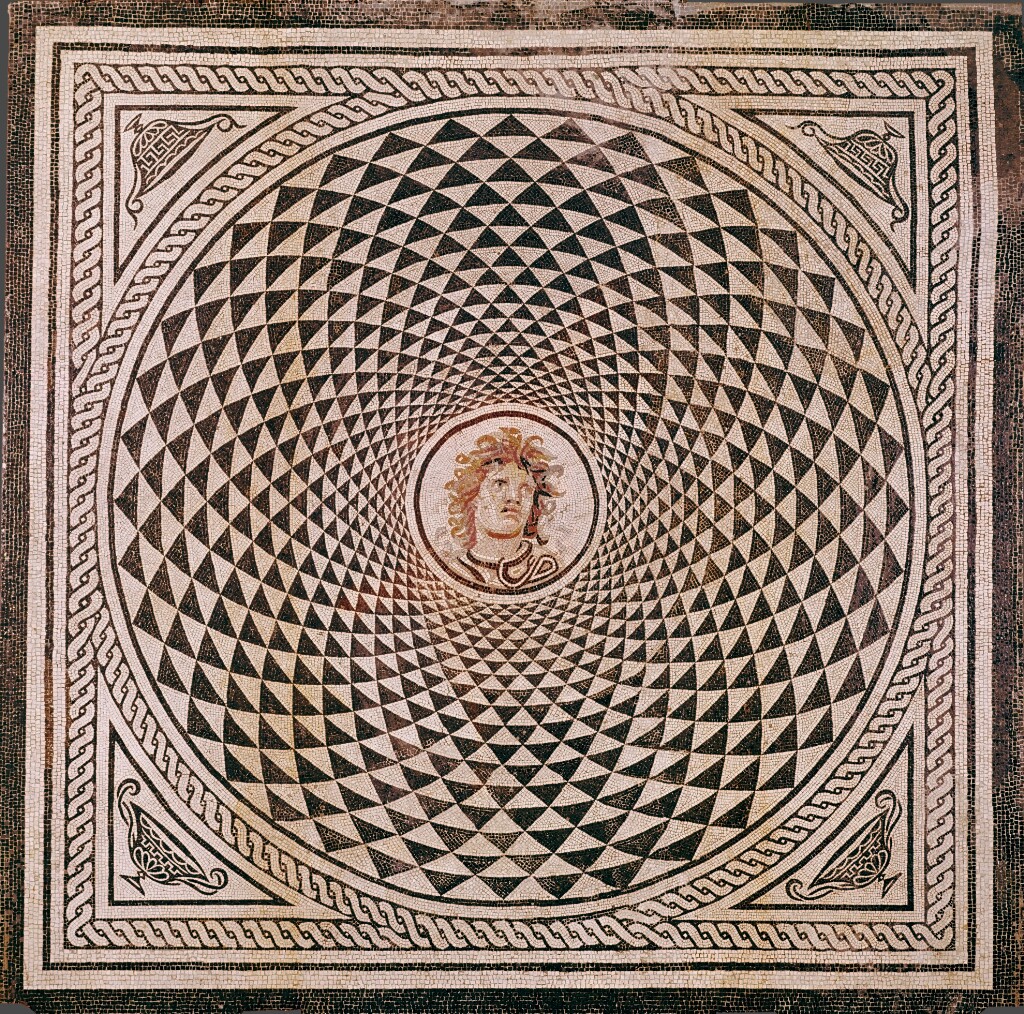}\\
      .\\
      \includegraphics[scale=0.19]{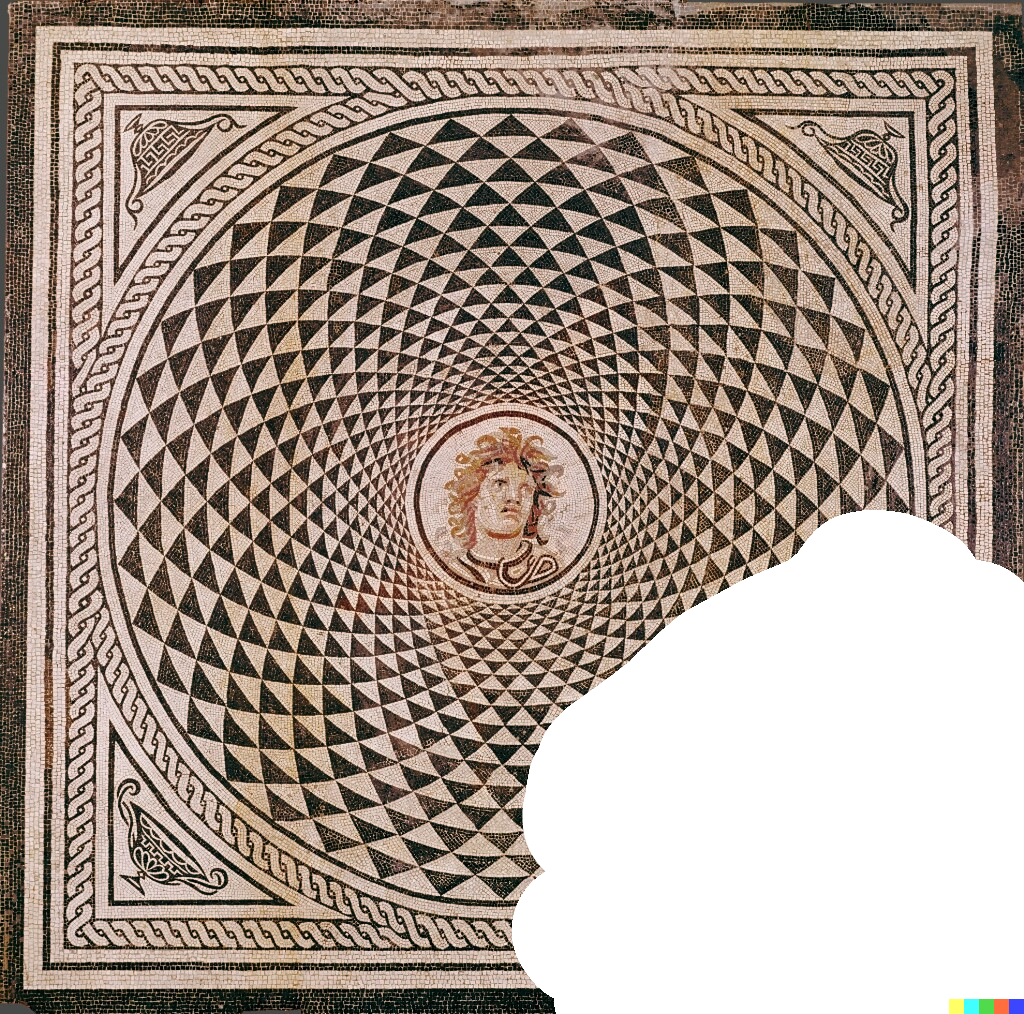}\\
      .\\
      \includegraphics[scale=0.19]{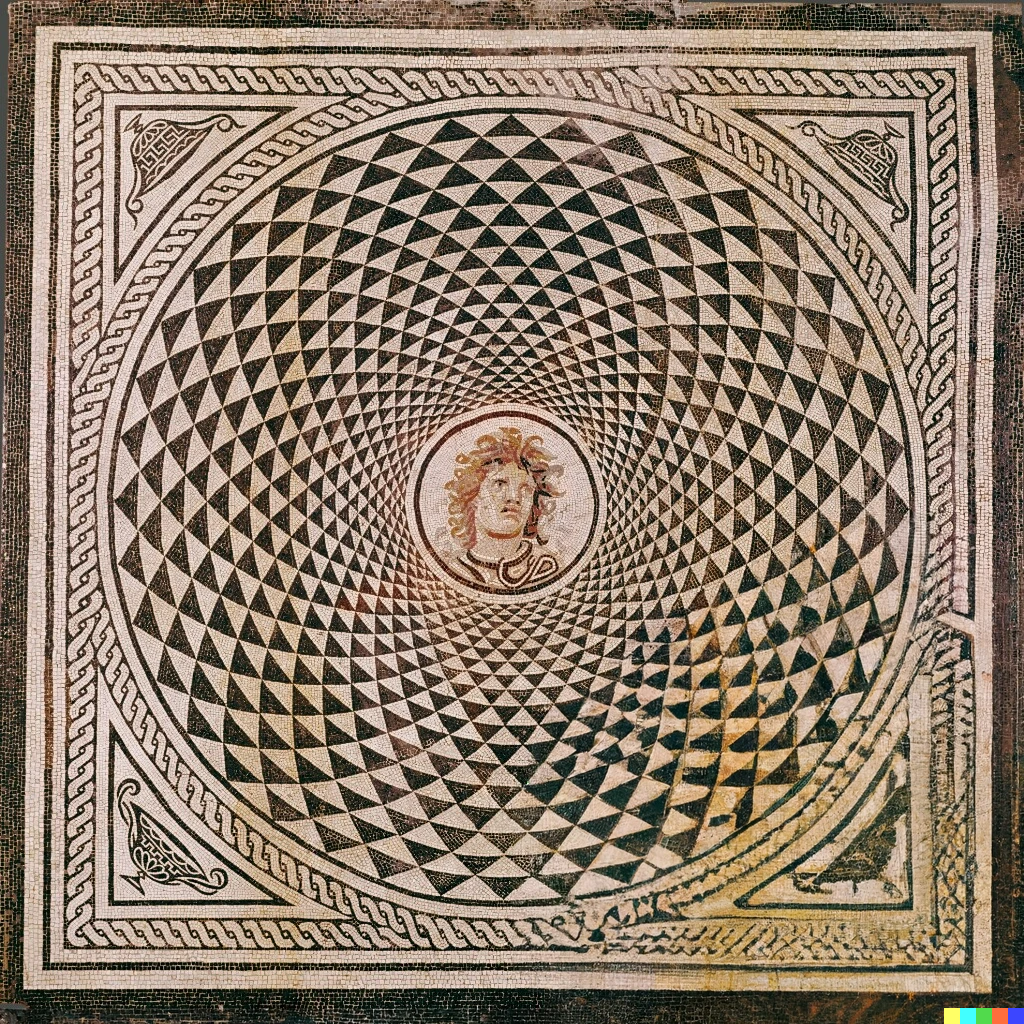}\\
    \caption{Mosaic Floor with Head of Medusa: original (top), emulated damage (middle), reconstructed (bottom)}
\end{figure*}

\subsection{Mosaic Floor with Orpheus and the Animals}

The last mosaic is a floor with Orpheus and the animals and the four seasons in the corners from around A.D. 150–200 that was found in France \footnote{Further details as well as the image with different resolutions are available at \url{https://www.getty.edu/art/collection/object/103QSM}}. Damaged has been introduced on selective locations across the mosaic.

The original image, the editing with the damage introduced artificially and the AI reconstructed image are shown in Figure 12. The AI tool is capable to reconstruct the scene correctly with good quality except for minor defects on the round decorative border and small geometric forms.

\begin{figure*}
   \label{FigM11}
    \centering
      \includegraphics[scale=0.38 ]{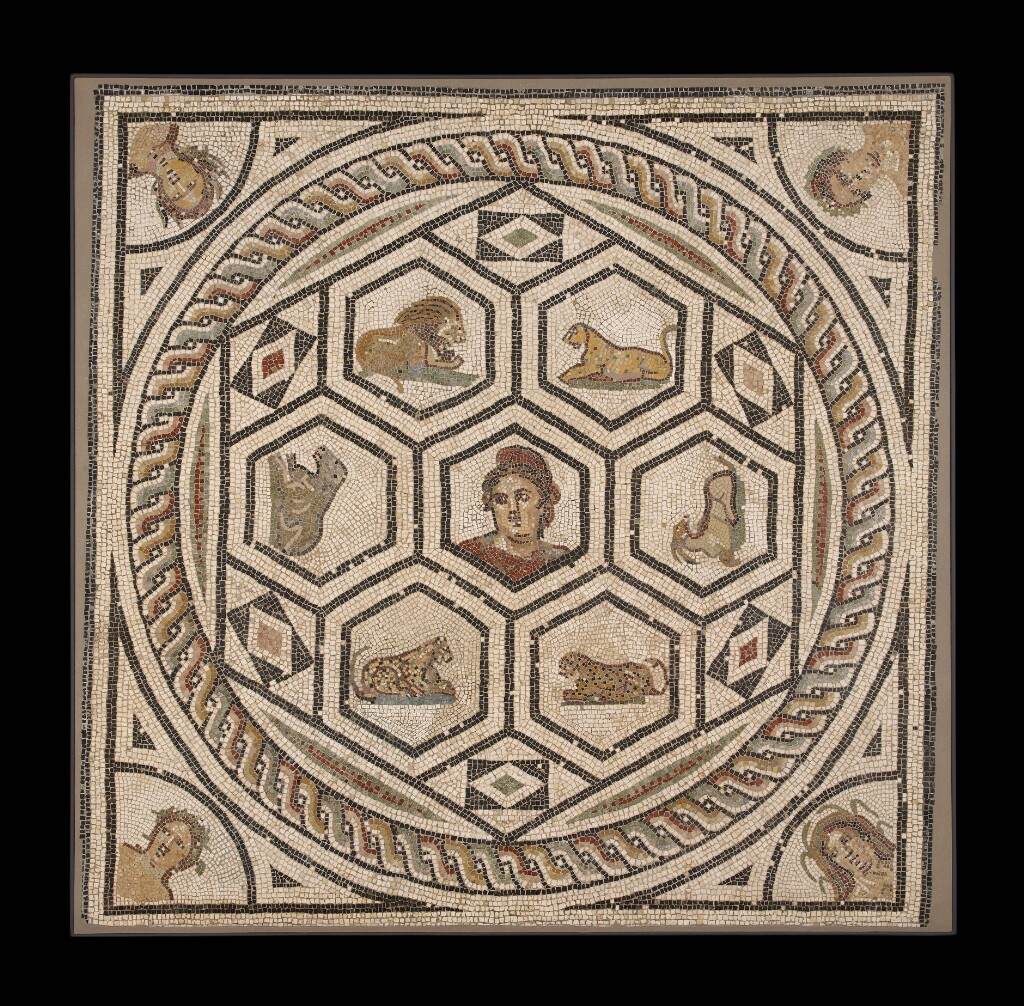}\\
      .\\
      \includegraphics[scale=0.18]{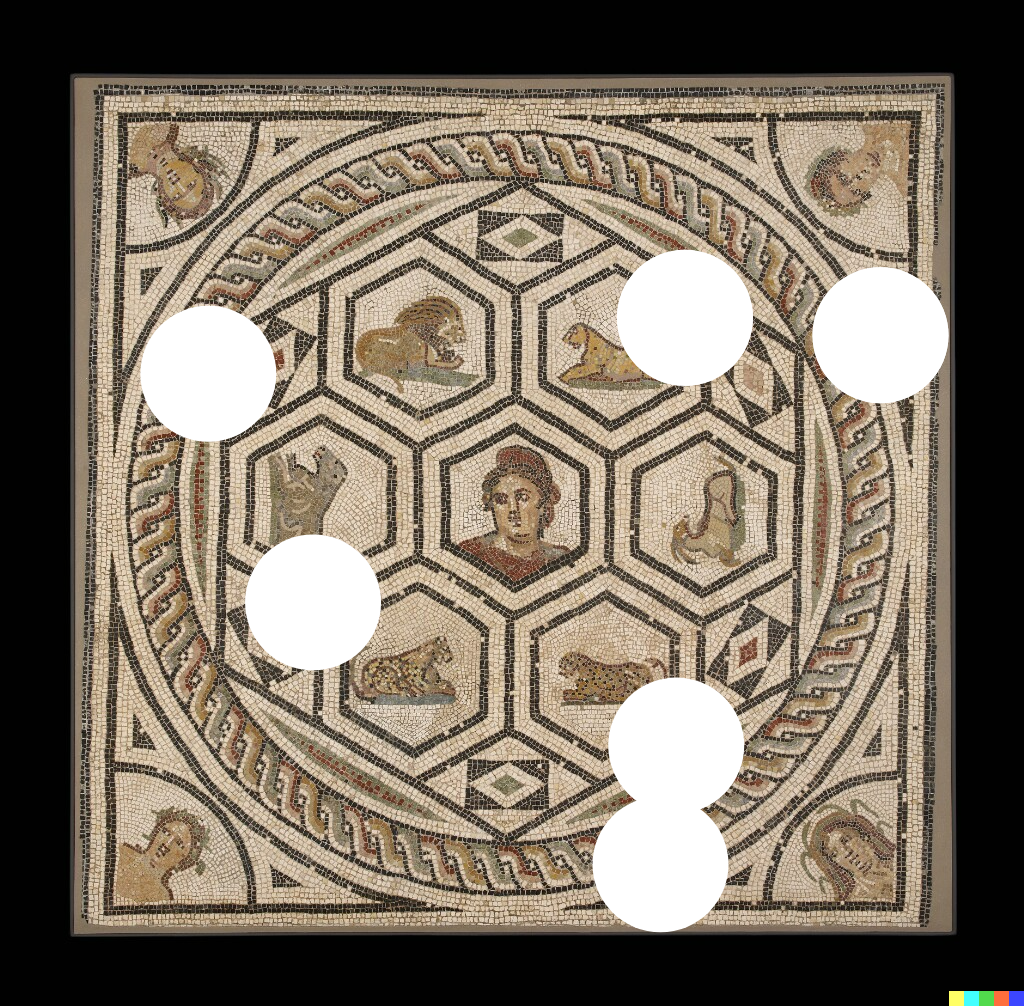}\\
      .\\
      \includegraphics[scale=0.18]{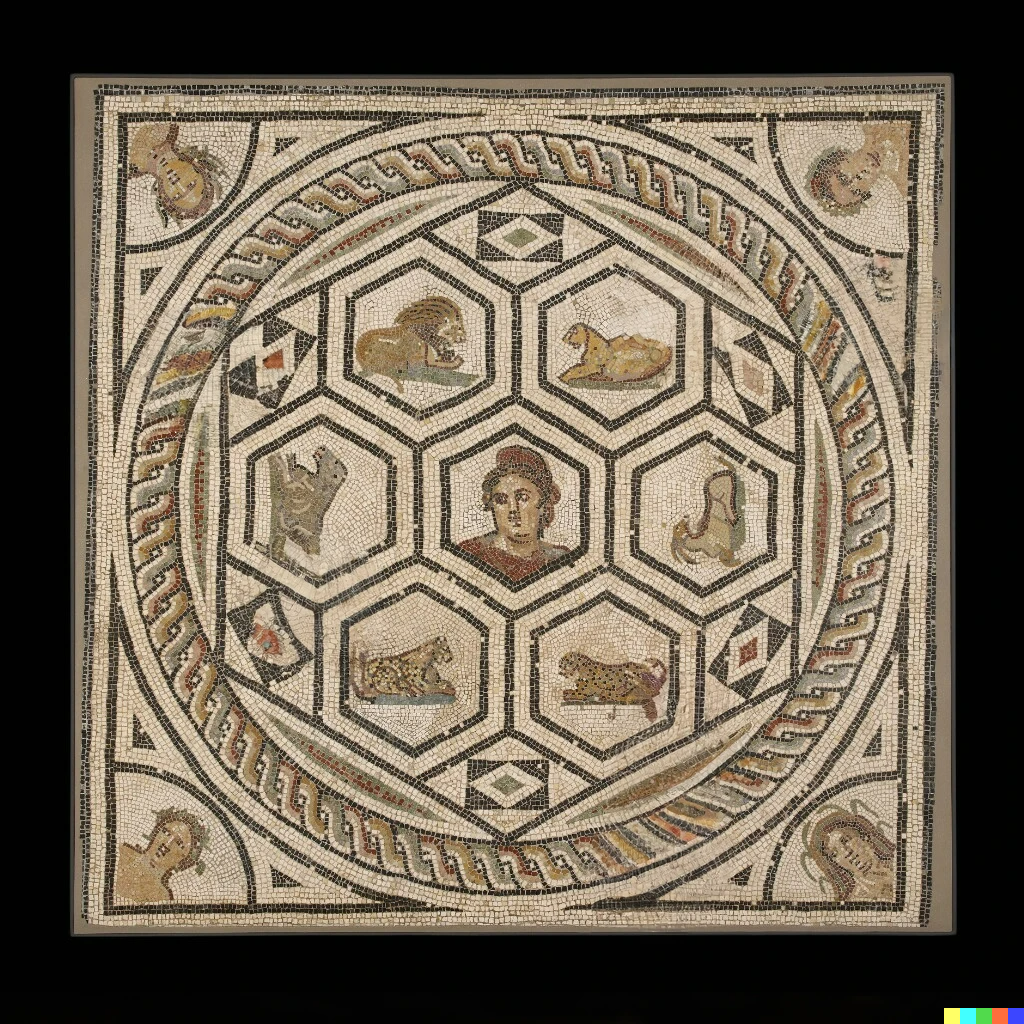}\\
    \caption{Mosaic Floor with Orpheus and the Animals: original (top), emulated damage (middle), reconstructed (bottom)}
\end{figure*}

\section{Analysis and discussion}
\label{sec:Discussion}

The evaluation conducted on different mosaics in the previous sections (either originally damaged or with simulated damage), show that DALL-E is capable to approximately interpret the main elements present on the scene and perform a meaningful reconstruction. This applies to scenes that contain persons and animals or to those with geometric forms and designs. However, the reconstructions show several mistakes and therefore, it is still far in most cases from the quality of a manual reconstruction. The worst performance is obtained when recreating faces and in the presence of nudity (this is due to DALL-E policies on the contents of images). For geometric shapes, the performance seems to be better, but DALL-E has some limitations on the color recreation and for some of the forms, especially when they are small. The tool has also limitations when having to reconstruct large missing parts for which the style and type of mosaic provide fundamental information. Even with these limitations, the performance of DALL-E is overall noteworthy if we take into consideration that the evaluation has been done on the Beta version of the outpainting feature using a very generic text prompt. 

Looking forward, we should expect major improvements in AI tool performance for mosaic reconstruction. DALL-E has continually implemented improvements with a second release of the tool and the addition of features such as outpainting in less than two years. The same applies to other companies developing similar tools, like Google with Imagen that was shortly followed by Parti or MidJourney that has released several versions of its tool. With the race to improve technology continuing and accelerating, we can only expect performance to improve. As this happens, AI image generators may become a valuable tool for the preservation of cultural heritage and will likely find new uses for cultural heritage reconstruction beyond mosaics.  

The performance of the AI tools for image generation and reconstruction will almost certainly improve as larger data sets and models are used and better algorithms are designed; new tools and updated versions of existing tools are being developed and introduced in the commercial market. These advancements will apply to nearly all types of images as this technology becomes ubiquitous in the coming years. However, specific improvements for mosaic reconstruction (and more generally for the reconstruction of art) can also be made; for example, a tailored AI tool can be trained only with mosaics and related images and texts for a tool that can interpret and reconstruct the details of the mosaics and also it makes it consistently with the commonly used patterns in the type of mosaic being reconstructed. The ranges of colors could also be trained. For example, there are mosaics with a generally more restricted palette which are made only with natural stones, while others, formed by mixtures of ceramics or tinted glasses, have much more variegated colors.  

A more subtle potential benefit of these tailored AI tools is that existing general tools seem to add a patina of time to the reconstruction. This is probably linked to the tools being trained with old images that show mosaics that have endured the passing of time. This also reveals the need to handle variables such as time that are beyond geometry and iconography. Taking the argument further, a central paving mosaic does not have the same wear as a lateral one; also a mosaic that has suffered an "accident" such as a fire or an earthquake does not have the same surface-distorting patina as the same but "preserved" by different layers of earth. These variables can lead to errors in recreations as encountered and commented on the analyzed images (and this does not depend on the good work of the original authors). By developing tailored AI tools, we could train the system only with images that have not suffered degradation to improve the quality of the recreations or even not only reconstructing the missing parts but removing the effects of the time on the entire mosaic. Similarly, it would be interesting that future AI tools could provide an indication of their confidence in the reconstruction (such as AI classifiers when performing classification). This would help to better interpret and use the AI generated images.

Another interesting topic for future research is to explore new applications of AI based image reconstruction for cultural heritage preservation. A third dimension could also be considered in these new avenues of research, because the possibilities in the fields of Architecture and Sculpture are wide-ranging. Some candidates could be not only damaged artifacts but also (purposed or not) unfinished ones, such as the slaves by Michelangelo or his Pietà Rondanini or worldwide known uncompleted architectures such as the Greek Temple of Segesta or the Malatesta Temple of Rimini. 

\section{Conclusion and future work}
\label{sec:Conclusion}

In this paper we have for the first time proposed and evaluated the use of Artificial Intelligence (AI) to reconstruct damaged ancient mosaics. The results show that state of the art AI tools such as DALL-E are capable of producing meaningful reconstructions of damaged mosaics. The current tool has still many limitations that result in reconstructions still far from the quality of a manual reconstruction. However, it is expected that performance of the AI tools will improve dramatically in the next few years opening new paths to recover and preserve mosaics. Looking further into the future, AI tools could be used to address challenges in other areas of cultural heritage preservation such as the reconstruction of three dimensional objects (like sculptures and buildings) and also the completion of unfinished work. As AI image generation technology keeps improving in the coming years we expect it to be a key enabler for innovation in cultural heritage research.

\section*{Acknowledgements}

This work was supported by the "Art, Architecture and Heritage in the processes of construction of the image of the new cultural enclaves (from the District to the Territory)" project PGC2018-094351-B-C43 and by the ACHILLES project PID2019-104207RB-I00 funded by the Spanish Agencia Estatal de Investigaci\'on (AEI) 10.13039/501100011033.




\bibliographystyle{IEEEtran}


\bibliography{mosaics}

\begin{thebibliography}{10}
\providecommand{\url}[1]{#1}
\csname url@samestyle\endcsname
\providecommand{\newblock}{\relax}
\providecommand{\bibinfo}[2]{#2}
\providecommand{\BIBentrySTDinterwordspacing}{\spaceskip=0pt\relax}
\providecommand{\BIBentryALTinterwordstretchfactor}{4}
\providecommand{\BIBentryALTinterwordspacing}{\spaceskip=\fontdimen2\font plus
\BIBentryALTinterwordstretchfactor\fontdimen3\font minus
  \fontdimen4\font\relax}
\providecommand{\BIBforeignlanguage}[2]{{%
\expandafter\ifx\csname l@#1\endcsname\relax
\typeout{** WARNING: IEEEtran.bst: No hyphenation pattern has been}%
\typeout{** loaded for the language `#1'. Using the pattern for}%
\typeout{** the default language instead.}%
\else
\language=\csname l@#1\endcsname
\fi
#2}}
\providecommand{\BIBdecl}{\relax}
\BIBdecl

\bibitem{Earthquakes}
\BIBentryALTinterwordspacing
L.~Alterio, G.~Russo, and F.~Silvestri, ``Seismic vulnerability reduction for
  historical buildings with non-invasive subsoil treatments: The case study of
  the mosaics palace at herculaneum,'' \emph{International Journal of
  Architectural Heritage}, vol.~11, no.~3, pp. 382--398, 2017. [Online].
  Available: \url{https://doi.org/10.1080/15583058.2016.1238969}
\BIBentrySTDinterwordspacing

\bibitem{virtualrestoration}
D.~Riccio, S.~Caggiano, M.~De~Marsico, R.~Distasi, and M.~Nappi, ``Mosaic+:
  tools to assist virtual restoration,'' 08 2015.

\bibitem{digitalimagingcultural}
F.~Stanco, S.~Battiato, and G.~Gallo, ``Digital imaging for cultural heritage
  preservation,'' \emph{Analysis, Restoration, and Reconstruction of Ancient
  Artworks}, 2011.

\bibitem{DigitalMosaicReconstructions1}
L.~Fazio, M.~Lo~Brutto, and G.~Dardanelli, ``Survey and virtual reconstruction
  of ancient roman floors in an archaeological context,'' \emph{ISPRS -
  International Archives of the Photogrammetry, Remote Sensing and Spatial
  Information Sciences}, vol. XLII-2/W11, pp. 511--518, 05 2019.

\bibitem{DigitalMosaicReconstructions2}
\BIBentryALTinterwordspacing
M.~Monti and G.~Maino, ``Non-metric digital reconstruction of roman mosaics
  excavated in the city of ravenna (italy),'' \emph{Virtual Archaeology
  Review}, vol.~9, no.~19, p. 66–75, Jul. 2018. [Online]. Available:
  \url{http://ojs.upv.es/index.php/var/article/view/7227}
\BIBentrySTDinterwordspacing

\bibitem{CulturalAI}
L.~Bordoni, F.~Mele, A.~Sorgente, P.~Mulholland, A.~Wolff, E.~Kilfeather, and
  M.~Maguire, ``Artificial intelligence for cultural heritage.''

\bibitem{AI-BIM}
\BIBentryALTinterwordspacing
D.~Bienvenido-Huertas, J.~E. Nieto-Julián, J.~J. Moyano, J.~M. Macías-Bernal,
  and J.~Castro, ``Implementing artificial intelligence in h-bim using the j48
  algorithm to manage historic buildings,'' \emph{International Journal of
  Architectural Heritage}, vol.~14, no.~8, pp. 1148--1160, 2020. [Online].
  Available: \url{https://doi.org/10.1080/15583058.2019.1589602}
\BIBentrySTDinterwordspacing

\bibitem{ImageSegmentationforMosaics}
A.~Bartoli, G.~Fenu, E.~Medvet, F.~A. Pellegrino, and N.~Timeus, ``Segmentation
  of mosaic images based on deformable models using genetic algorithms,'' in
  \emph{Smart Objects and Technologies for Social Good}, O.~Gaggi, P.~Manzoni,
  C.~Palazzi, A.~Bujari, and J.~M. Marquez-Barja, Eds.\hskip 1em plus 0.5em
  minus 0.4em\relax Cham: Springer International Publishing, 2017, pp.
  233--242.

\bibitem{ImageSegmentationML}
A.~Felicetti, M.~Paolanti, P.~Zingaretti, R.~Pierdicca, and E.~S. Malinverni,
  ``Mo. se.: Mosaic image segmentation based on deep cascading learning,''
  \emph{Virtual Archaeology Review}, vol.~12, no.~24, pp. 25--38, 2021.

\bibitem{Imagen}
\BIBentryALTinterwordspacing
C.~Saharia, W.~Chan, S.~Saxena, L.~Li, J.~Whang, E.~Denton, S.~K.~S.
  Ghasemipour, B.~K. Ayan, S.~S. Mahdavi, R.~G. Lopes, T.~Salimans, J.~Ho,
  D.~J. Fleet, and M.~Norouzi, ``Photorealistic text-to-image diffusion models
  with deep language understanding,'' 2022. [Online]. Available:
  \url{https://arxiv.org/abs/2205.11487}
\BIBentrySTDinterwordspacing

\bibitem{Parti}
\BIBentryALTinterwordspacing
J.~Yu, Y.~Xu, J.~Y. Koh, T.~Luong, G.~Baid, Z.~Wang, V.~Vasudevan, A.~Ku,
  Y.~Yang, B.~K. Ayan, B.~Hutchinson, W.~Han, Z.~Parekh, X.~Li, H.~Zhang,
  J.~Baldridge, and Y.~Wu, ``Scaling autoregressive models for content-rich
  text-to-image generation,'' 2022. [Online]. Available:
  \url{https://arxiv.org/abs/2206.10789}
\BIBentrySTDinterwordspacing

\bibitem{Cogview}
\BIBentryALTinterwordspacing
M.~Ding, Z.~Yang, W.~Hong, W.~Zheng, C.~Zhou, D.~Yin, J.~Lin, X.~Zou, Z.~Shao,
  H.~Yang, and J.~Tang, ``Cogview: Mastering text-to-image generation via
  transformers,'' 2021. [Online]. Available:
  \url{https://arxiv.org/abs/2105.13290}
\BIBentrySTDinterwordspacing

\bibitem{DALLE1}
\BIBentryALTinterwordspacing
A.~Ramesh, M.~Pavlov, G.~Goh, S.~Gray, C.~Voss, A.~Radford, M.~Chen, and
  I.~Sutskever, ``Zero-shot text-to-image generation,'' 2021. [Online].
  Available: \url{https://arxiv.org/abs/2102.12092}
\BIBentrySTDinterwordspacing

\bibitem{DALLE2}
\BIBentryALTinterwordspacing
A.~Ramesh, P.~Dhariwal, A.~Nichol, C.~Chu, and M.~Chen, ``Hierarchical
  text-conditional image generation with clip latents,'' 2022. [Online].
  Available: \url{https://arxiv.org/abs/2204.06125}
\BIBentrySTDinterwordspacing

\bibitem{HandbookRomanArt}
\emph{\BIBforeignlanguage{und}{A handbook of Roman art : a survey of the visual
  arts of the Roman world}}.\hskip 1em plus 0.5em minus 0.4em\relax Oxford:
  Phaidon, 1983.

\bibitem{Thruxton}
\BIBentryALTinterwordspacing
M.~Henig and G.~Soffe, ``The thruxton roman villa and its mosaic pavement,''
  \emph{Journal of the British Archaeological Association}, vol. 146, no.~1,
  pp. 1--28, 1993. [Online]. Available:
  \url{https://doi.org/10.1179/jba.1993.146.1.1}
\BIBentrySTDinterwordspacing

\bibitem{Gestaltprinciples}
K.~Koffka, \emph{Principles of Gestalt psychology}.\hskip 1em plus 0.5em minus
  0.4em\relax Routledge, 2013.

\bibitem{GestaltCNN}
A.~Amanatiadis, V.~G. Kaburlasos, and E.~B. Kosmatopoulos, ``Understanding deep
  convolutional networks through gestalt theory,'' in \emph{2018 IEEE
  International Conference on Imaging Systems and Techniques (IST)}, 2018, pp.
  1--6.

\end{thebibliography}

\newpage

\end{document}